\documentclass[10pt,twocolumn,letterpaper]{article}
 \usepackage{multirow}
\usepackage{cvpr}              
\definecolor{cvprblue}{rgb}{0.21,0.49,0.74}
\usepackage[pagebackref,breaklinks,colorlinks,allcolors=cvprblue]{hyperref}
\usepackage{algorithm}
\usepackage{algpseudocode}
\algrenewcommand{\algorithmicrequire}{\textbf{Input:}}
\let\Input\Require
\def\paperID{29700} 
\def\confName{CVPR}
\def\confYear{2026}
\newcommand{\model}{World-Env}
\title{\model{}: Leveraging World Model as a Virtual Environment for VLA Post-Training}
\newcommand\xiao[1]{\textcolor{black}{#1}}

\author{
   Junjin Xiao$^{1,2,*}$\quad  
   Yandan Yang$^{2}$\quad  
   Xinyuan Chang$^{2}$\quad
   Ronghan Chen$^{2}$ \\
   Feng Xiong$^{2}$\quad
   Mu Xu$^{2}$ \quad
   Wei-Shi Zheng$^{1,3}$ \quad
   Qing Zhang$^{1,3,\dagger}$\\
   \small $^1$School of Computer Science and Engineering, Sun Yat-sen University, China \\
  \small $^2$AMap, Alibaba Group\quad
   \small $^3$Key Laboratory of Machine Intelligence and Advanced Computing, Ministry of Education, China\\
}
\begin{document}
\twocolumn[{%
\renewcommand\twocolumn[1][]{#1}%
\maketitle
\includegraphics[width=1\linewidth]{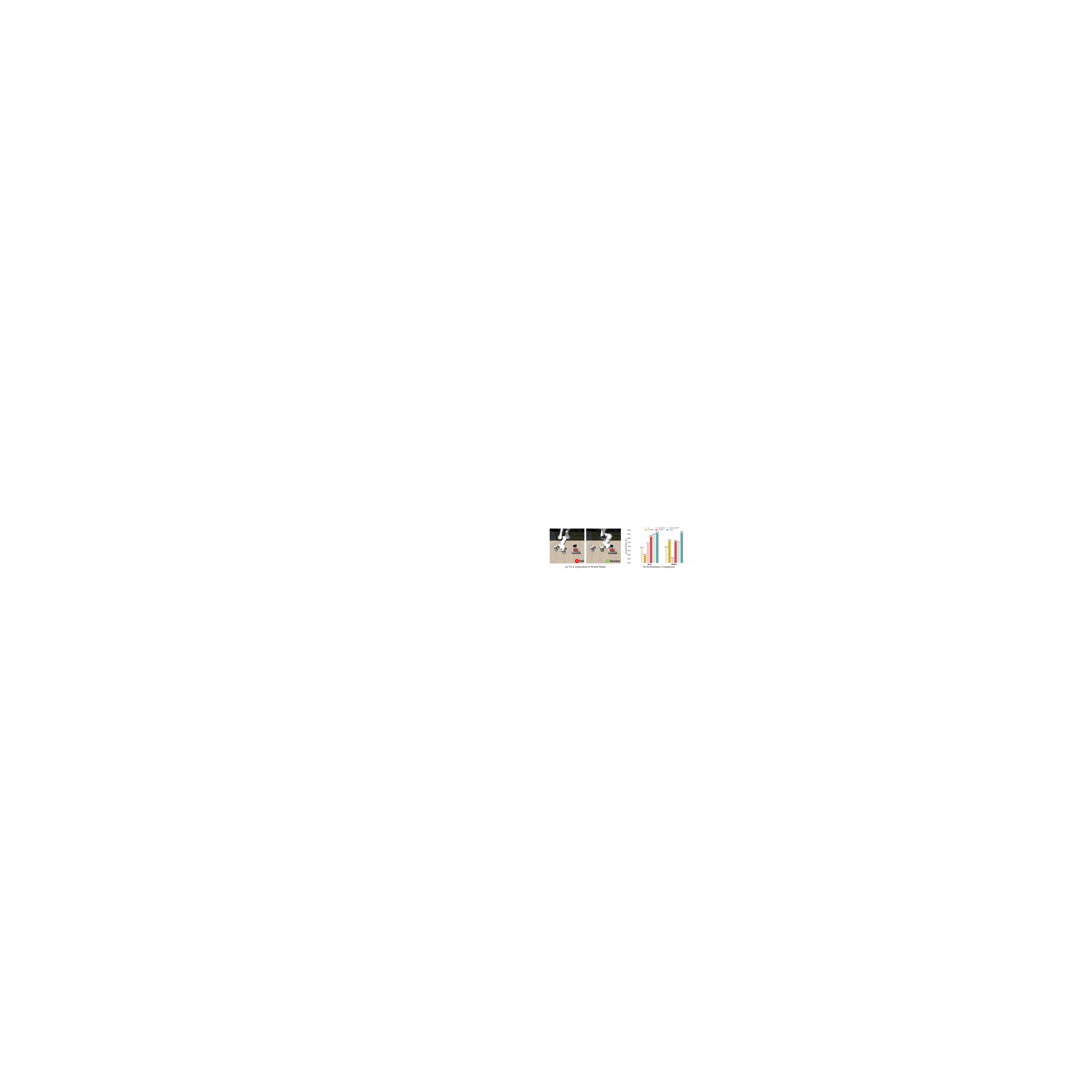}
\captionof{figure}{\textbf{VLA exploration in world model and performance comparison on LIBERO benchmarks.} }
\vspace{4mm}
\label{fig:teaser}
}]

\renewcommand{\thefootnote}{}
\footnotetext{$^*$Work done during an internship with AMap, Alibaba Group. $^\dagger$Corresponding author (zhangq93@mail.sysu.edu.cn).}

\begin{abstract}
Vision-Language-Action (VLA) models trained via imitation learning suffer from significant performance degradation in data-scarce scenarios due to their reliance on large-scale demonstration datasets. Although reinforcement learning (RL)-based post-training has proven effective in addressing data scarcity, its application to VLA models is hindered by the non-resettable nature of real-world environments. This limitation is particularly critical in high-risk domains such as industrial automation, where interactions often induce state changes that are costly or infeasible to revert. Furthermore, existing VLA approaches lack a reliable mechanism for detecting task completion, leading to redundant actions that reduce overall task success rates. To address these challenges, we propose \model{}, an RL-based post-training framework that replaces physical interaction with a low-cost world model-based virtual simulator. \model{} consists of two key components: (1) a \xiao{physically-consistent} world simulator that generates temporally consistent future visual observations, and (2) a vision-language model (VLM)-guided instant reflector that provides continuous reward signals and predicts action termination. This simulated environment enables VLA models to safely explore and generalize beyond their initial imitation learning distribution. Our method achieves notable performance gains with as few as five expert demonstrations per task. Experiments on complex robotic manipulation tasks demonstrate that \model{} effectively overcomes the data inefficiency, safety constraints, and inefficient execution of conventional VLA models that rely on real-world interaction, offering a practical and scalable solution for post-training in resource-constrained settings. 
Our code is available at \url{https://github.com/amap-cvlab/world-env}.

\end{abstract}    
\section{Introduction}
\label{sec:intro}
Vision-Language-Action (VLA) models have emerged as a central paradigm for autonomous agents, enabling end-to-end mapping from high-level language instructions to low-level motor commands by integrating vision, language, and control. These models have demonstrated considerable promise in robotic manipulation \cite{kim24openvla,black2024pi0visionlanguageactionflowmodel}, autonomous driving \cite{Yurtsever2020AutonomousDriving,liang2025persistent,yuan2025unimapgen,wan2025driving,zeng2025FSDrive}, and navigation \cite{Hong_2021_CVPR,zeng2025janusvln}. Most existing approaches rely on supervised fine-tuning through imitation learning, building upon pre-trained vision-language models \cite{touvron2023llamaopenefficientfoundation} to align semantic intent with physical execution via cross-modal representations.

However, imitation learning methods \cite{kim2025openvla-oft} are inherently constrained by the limited availability of high-quality demonstrations. In many real-world scenarios, collecting diverse and safe human demonstrations is prohibitively expensive and often infeasible due to safety concerns and environmental complexity. Furthermore, such methods generalize poorly to novel tasks or unseen objects, and their performance degrades under few-shot conditions. 

To overcome these shortcomings, recent works \cite{tan2025interactiveposttrainingvisionlanguageactionmodels,lu2025vlarlmasterfulgeneralrobotic} have turned to reinforcement learning (RL) \cite{rafailov2023dpo} to enable agents to learn through interaction. Current RL strategies fall into two categories. The first involves real-world learning with human feedback, which captures realistic environmental dynamics but suffers from non-resettable interactions, high trial costs, and limited reproducibility, rendering it unsuitable for safety-critical applications. The second relies on simulator-based learning, which avoids physical risks but introduces other challenges, including substantial development effort, limited sim-to-real transfer, and difficulty adapting to new objects or dynamic scene changes, thereby restricting its practical applicability.

\begin{figure}[t]
\begin{center}
\includegraphics[width=1\linewidth]{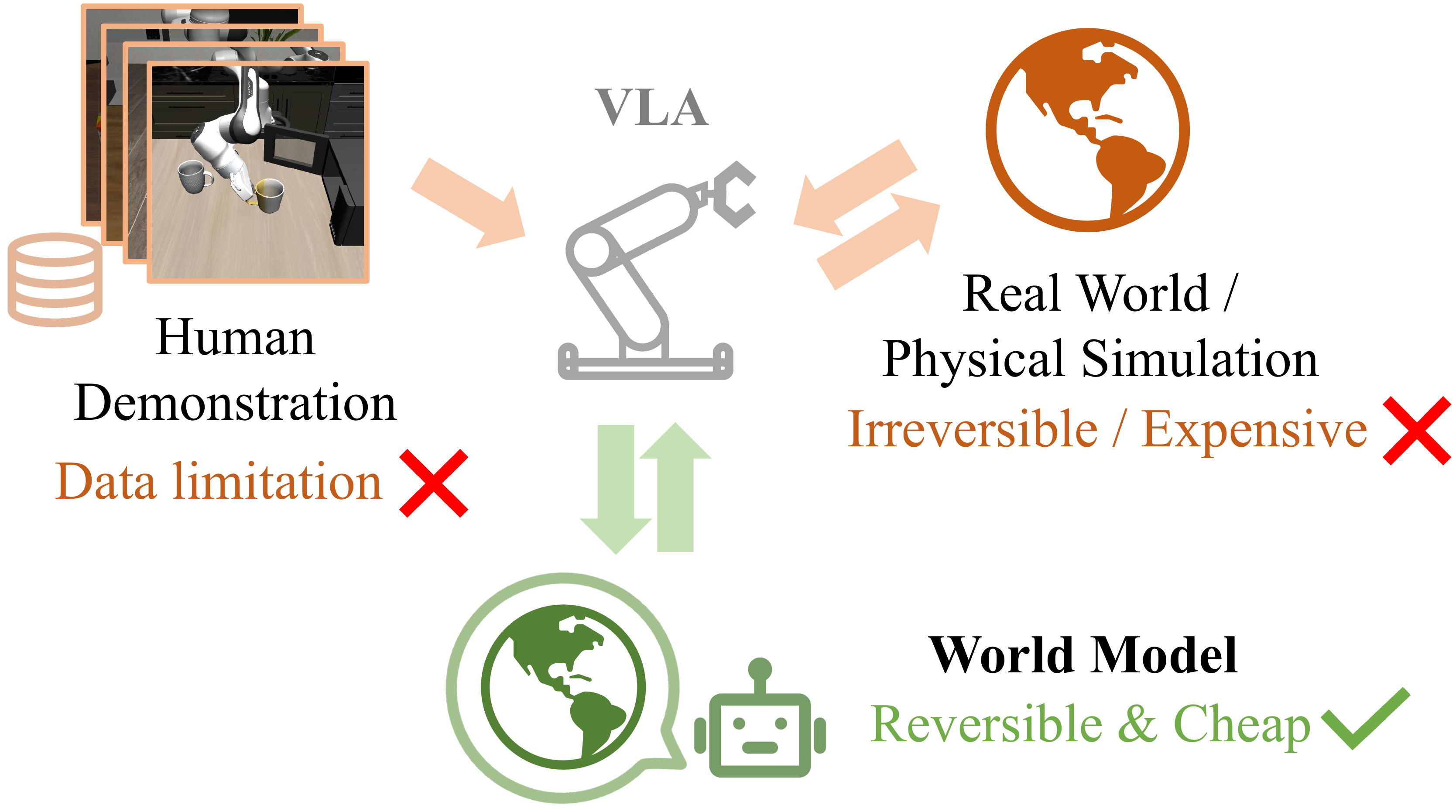}
\end{center}
\vspace{-5mm}
\caption{\textbf{Comparison of three VLA training paradigms.}}
\label{fig:teaser2}
\end{figure}

These limitations motivate us to think about a question: \textit{Is there an ``ideal testbed'' that avoids real-world risks while providing greater flexibility and richer semantic understanding than conventional simulators?} We find that video-based world model offers a promising solution. Equipped with action-conditioned future prediction and a persistent latent scene representation, world model can generate visually plausible future image sequences, allowing safe, low-cost simulation of action outcomes, as well as policy exploration and refinement without physical interaction.

In this work, we introduce \model{}, a world model-based reinforcement learning framework that improves policy generalization under data scarcity while respecting real-world safety constraints, as shown in Figure~\ref{fig:teaser2}. \model{} consists of two components. The first is a \xiao{physically-consistent} world simulator that functions as an interactive future-frame predictor, synthesizing action-conditioned image sequences that capture post-interaction object states and surrounding scene structure. \xiao{To ensure physical consistent of our world model, we propose a geometry-aware feature injection strategy, which leverages latent feature from VGGT~\cite{wang2025vggt} as additional condition.} The second is a VLM-guided instant reflector that functions as a semantics-aware reward module. It provides continuous reward signals by evaluating the semantic alignment between predicted visual frames and the input language instruction. This assessment supports policy optimization and enables real-time detection of task completion. Upon confirming successful execution (e.g., when the goal state is reached), the reflector immediately terminates the action sequence to prevent redundant or disruptive subsequent actions. 

In summary, our contributions are:

\begin{itemize}[leftmargin=2em] 
\setlength\itemsep{0.5em}
\item We propose \model{}, a framework that enables low-cost, safe reinforcement learning post-training for VLA policies under extreme data scarcity, eliminating the need for real-world interaction.
\item \xiao{We propose a geometry-aware feature injection strategy via injecting latent feature from VGGT to ensure physical consistent of world model.}
\item We introduce a real-time termination mechanism via the instant reflector, which dynamically assesses task completion by evaluating semantic alignment between predicted visual trajectories and language instructions, thereby preventing redundant post-success actions.
\end{itemize}

\section{Related Work}
\label{sec:related}
\begin{figure*}[t]
\begin{center}
\includegraphics[width=1\textwidth]{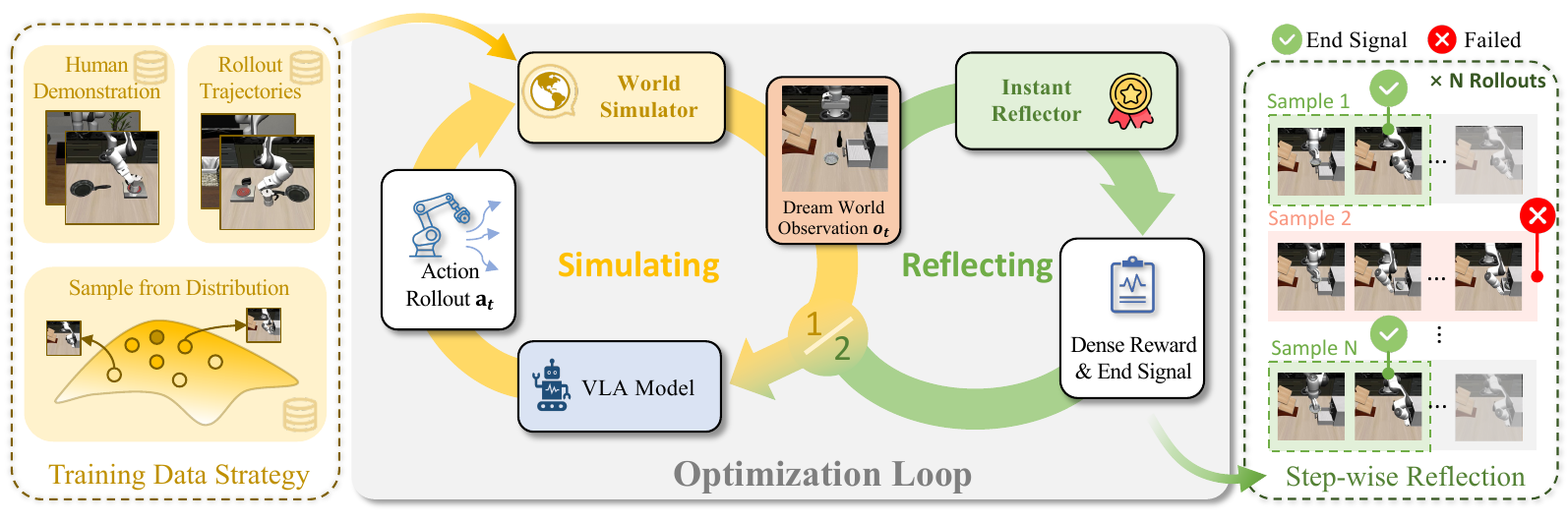} \\
\end{center}
\vspace{-8mm}
\caption{\textbf{Overview of \model{}.} Our framework comprises: (1) a \textit{Training Data Strategy} that augments human demonstrations trajectories with VLA self-explored trajectories to train the World Simulator; (2) an \textit{Optimization Loop} where the VLA model generates actions, the simulator predicts future observations, and the Instant Reflector generates feedback; and (3) \textit{Reward \& End Signal} provides trajectory-wise reward and end signals for RL optimization.}
\label{fig:pipeline}

\end{figure*}
\vspace{0.5em}
\noindent \textbf{Vision-Language-Action Models.} 
Leveraging advancements in pre-trained vision foundation models \cite{radford2021CLIP,oquab2023dinov2,dosovitskiy2020vit}, large language models (LLMs) \cite{brown2020gpt3,touvron2023llama,bai2023qwentechnicalreport}, and vision-language models (VLMs) \cite{alayrac2022flamingo,li2022blip,li2023blip2,dai2023instructblip,liu2023llava,liu2023improvedllava}, Vision-Language-Action (VLA) frameworks \cite{Chi2023DiffusionPolicy,octomodelteam2024octoopensourcegeneralistrobot,kim24openvla,bu2025univla} have emerged as a powerful approach for embodied intelligence. Specifically, DiffusionPolicy \cite{Chi2023DiffusionPolicy} proposes a diffusion-based policy that generates robot actions through a conditional denoising diffusion process in the action space, iteratively refining actions based on visual observations. OpenVLA \cite{kim24openvla} integrates robotic actions into a language modeling framework by mapping action sequences to discrete tokens within a large language model. The recent work OpenVLA-OFT \cite{kim2025openvla-oft} further converts discrete action sequences into continuous representations, achieving improved efficiency and performance.

\vspace{0.5em}
\noindent \textbf{Reinforcement Learning for VLA Systems.} 
Recent advances in reinforcement learning (RL) \cite{schulman2017proximal,rafailov2023dpo} have demonstrated considerable potential in enhancing decision-making capabilities of large language models (LLMs) \cite{guo2025deepseek,lightman2023let,ouyang2022training,lee2023rlaif}. This progress has spurred growing interest in applying RL to Vision-Language-Action (VLA) systems \cite{tan2025interactiveposttrainingvisionlanguageactionmodels,lu2025vlarlmasterfulgeneralrobotic,chandra2025diwa,jiang2025irl}. Unlike supervised fine-tuning (SFT), RL enables agents to refine policies through interaction. This paradigm supports autonomous exploration and improved robustness to partial observability, allowing VLA models to generalize to unseen scenarios while reducing reliance on costly human demonstrations. However, existing RL-based VLA methods typically require real-world interaction, which is often infeasible in high-risk scenarios.

\vspace{0.5em}
\noindent \textbf{World Models.}
World models \cite{assran2025vjepa2,genie3}, which are learned dynamical simulators that approximate environmental dynamics, have become foundational for safe and sample-efficient agent training. Early works \cite{hafner2019learning,hafner2020mastering,hafner2025mastering} demonstrated the effectiveness of model-based reinforcement learning in virtual environments, enabling agents to plan via imagined trajectories without real-world interaction. However, these methods typically rely on on-policy data, limiting their generalization to specific environments and downstream tasks. Building on advances in diffusion-based video generation \cite{ho2020denoising,rombach2022high,blattmann2023stable,yang2024cogvideox,wan2025,xing2024dynamicrafter}, we propose a framework that trains a world model on offline demonstration data and keeps it fixed during policy learning to predict future visual observations for VLA models.

\section{Preliminary}
\vspace{0.5em}
\noindent \textbf{Vision-Language-Action Model.}
Vision-language-action (VLA) models bridge natural language instructions with robotic control by translating semantic goals into low-level actions while grounding language in multimodal observations. Following recent VLA frameworks such as OpenVLA-OFT \cite{kim2025openvla-oft}, the policy is implemented as a deterministic mapping that leverages a pretrained vision-language model to extract multimodal features, followed by a lightweight action head for continuous control. Specifically, given a history of RGB observations $\mathbf{o}_{1:t}$, proprioceptive states $\mathbf{s}_{1:t}$ (e.g., joint angles or end-effector poses), and a language instruction $\mathbf{g}$, the policy predicts a action as:
\begin{equation}
\mathbf{a}_t = \pi_{\theta}(\mathbf{o}_{1:t}, \mathbf{s}_{1:t}, \mathbf{g}),    
\end{equation}
where $\pi_{\theta}$ denotes a deterministic policy parameterized by a finetuned foundation model and a trainable action head.

\vspace{0.5em}
\noindent \textbf{Reinforcement Learning.}
Reinforcement learning (RL) formulates decision-making as a Markov Decision Process (MDP): $\mathcal{M} = (\mathcal{S}, \mathcal{A}, \mathcal{P}, \mathcal{R}, \gamma)$, where $\mathcal{S}$ is the state space (comprising visual observations $\mathbf{o}_t$ and proprioceptive states $\mathbf{s}_t$), $\mathcal{A}$ is the action space (e.g., continuous control commands $\mathbf{a}_t \in \mathbb{R}^D$), $\mathcal{P}$ denotes the transition dynamics, $\mathcal{R}$ is the reward function, and $\gamma \in [0,1]$ is the discount factor. The objective is to learn a policy $\pi_{\theta}(\mathbf{o}_{1:t}, \mathbf{s}_{1:t}, \mathbf{g})$ that maximizes the expected return:
\begin{equation}
    J(\pi_{\theta}) = \mathbb{E}_{\pi_{\theta}} \left[ \sum_{t=0}^T \gamma^t r_t \right],
\end{equation}
where $r_t = \mathcal{R}(\mathbf{o}_{1:t}, \mathbf{g})$. In practice, policy gradient methods often introduce stochasticity during training to enable exploration. The policy is updated using gradients:
\begin{equation}
    \nabla_\theta J(\pi_\theta) = \mathbb{E}_{\mathbf{a}_t \sim \pi_\theta } \left[ \nabla_\theta \log \pi_\theta(\mathbf{o}_{1:t}, \mathbf{s}_{1:t}, \mathbf{g}) \cdot A(\mathbf{o}_{1:t}, \mathbf{a}_{1:t}) \right],
\end{equation}
where $A(\cdot)$ is the advantage function that evaluates action quality relative to a baseline.
\section{Method}
\begin{figure}[t]
\begin{center}
\includegraphics[width=1\linewidth]{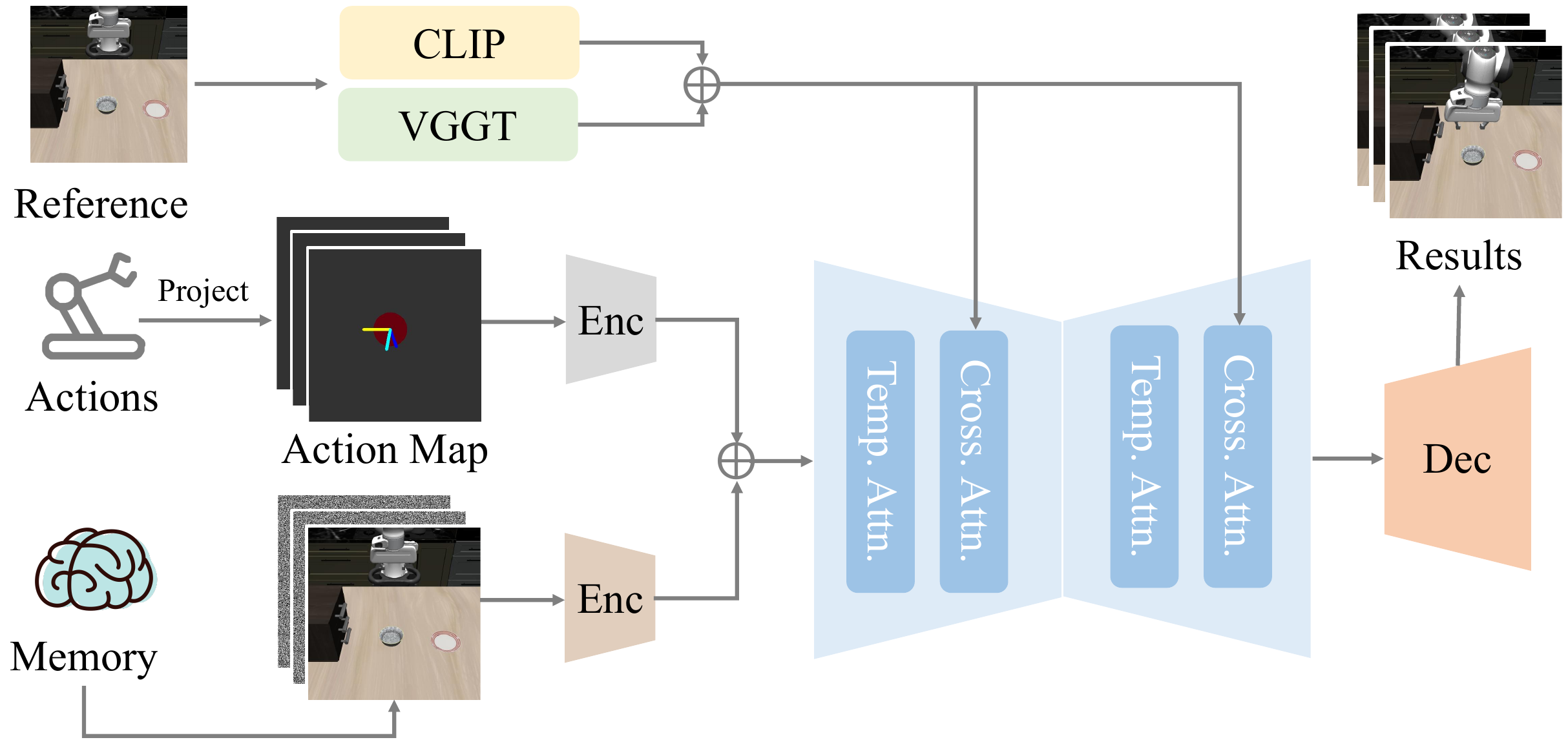}
\end{center}
\vspace{-5mm}
\caption{\textbf{Overview of physically-consistent world simulator.}}
\label{fig:wm_pipeline}
\end{figure}
Figure~\ref{fig:pipeline} presents the overview of our framework. Prior VLA approaches~\cite{kim24openvla,kim2025openvla-oft} typically rely on either real-world interaction or conventional simulators to provide observations for action prediction. In contrast, our framework eliminates the need for physical interaction by leveraging a \textit{physically-consistent world simulator} that generates temporally consistent future visual observations at low cost. Specifically, the VLA policy $\pi_\theta$ maps the current RGB observation $\mathbf{o}_t$, language instruction $\mathbf{g}$, and proprioceptive state $\mathbf{s}_t$ (comprising the 6D end-effector pose and 1D gripper state) to a continuous action $\mathbf{a}_t$. The next proprioceptive state $\mathbf{s}_{t+1}$ is then computed deterministically using forward kinematics. The world simulator takes the resulting proprioceptive state $\mathbf{s}_{t+1}$ as input and predicts the subsequent visual observation $\mathbf{o}_{t+1}$. This imagined observation, together with $\mathbf{s}_{t+1}$, is fed back into the VLA policy to predict the next action $\mathbf{a}_{t+1}$. The rollout terminates either when the maximum timestep is reached or when the \textit{VLM-guided instant reflector}, which evaluates semantic alignment between the predicted visual trajectory and the language instruction, confirms task success and issues a termination signal. During training, we collect $N$ simulated trajectories from this virtual environment and use them for RL optimization of the VLA policy within \model{}.

\begin{table*}[]
\centering
\caption{\textbf{Success rate comparison on the LIBERO benchmark.} We report success rates for each method using the same setting with only 5 demonstrations per task.}
\vspace{-2mm}
\resizebox{\linewidth}{!}
{
\begin{tabular}{l c c c c c}
\toprule[1pt]
\multirow{1}{*}{Method}& \multicolumn{1}{c}{LIBERO-Goal}& \multicolumn{1}{c}{LIBERO-Object}& \multicolumn{1}{c}{ LIBERO-Spatial}&\multicolumn{1}{c}{LIBERO-Long}&\multicolumn{1}{c}{Average}\\ 
\midrule
$\pi_0$~\cite{black2024pi0visionlanguageactionflowmodel}& 67.6&68.4 & 80.2& 28.2& 61.1 \\
$\pi_0$+FAST~\cite{pertsch2025fast}&59.2 & 76.8& 59.2& 24.8&  55.0\\
OpenVLA~\cite{kim24openvla}& 73.2 & 55.0 & 82.4 & 32.2 & 60.7 \\
UniVLA~\cite{bu2025univla}& \underline{82.0} & \underline{76.2} & \underline{84.4} & 56.4 &  74.75 \\
OpenVLA-OFT~\cite{kim2025openvla-oft}& 84.0 & 74.2 & 84.2 & \underline{57.0} & \underline{74.85}  \\
OpenVLA-OFT + Post-training (Ours)& \textbf{86.4} &\textbf{86.6} &\textbf{87.6} & \textbf{57.8} &  \textbf{79.6}\\
\bottomrule[1pt]
\end{tabular}
}
\label{table:main_compare}
\end{table*}

\begin{figure*}[t]
\centering
  \begin{minipage}{0.32\textwidth}
    \centering
    \includegraphics[width=\linewidth]{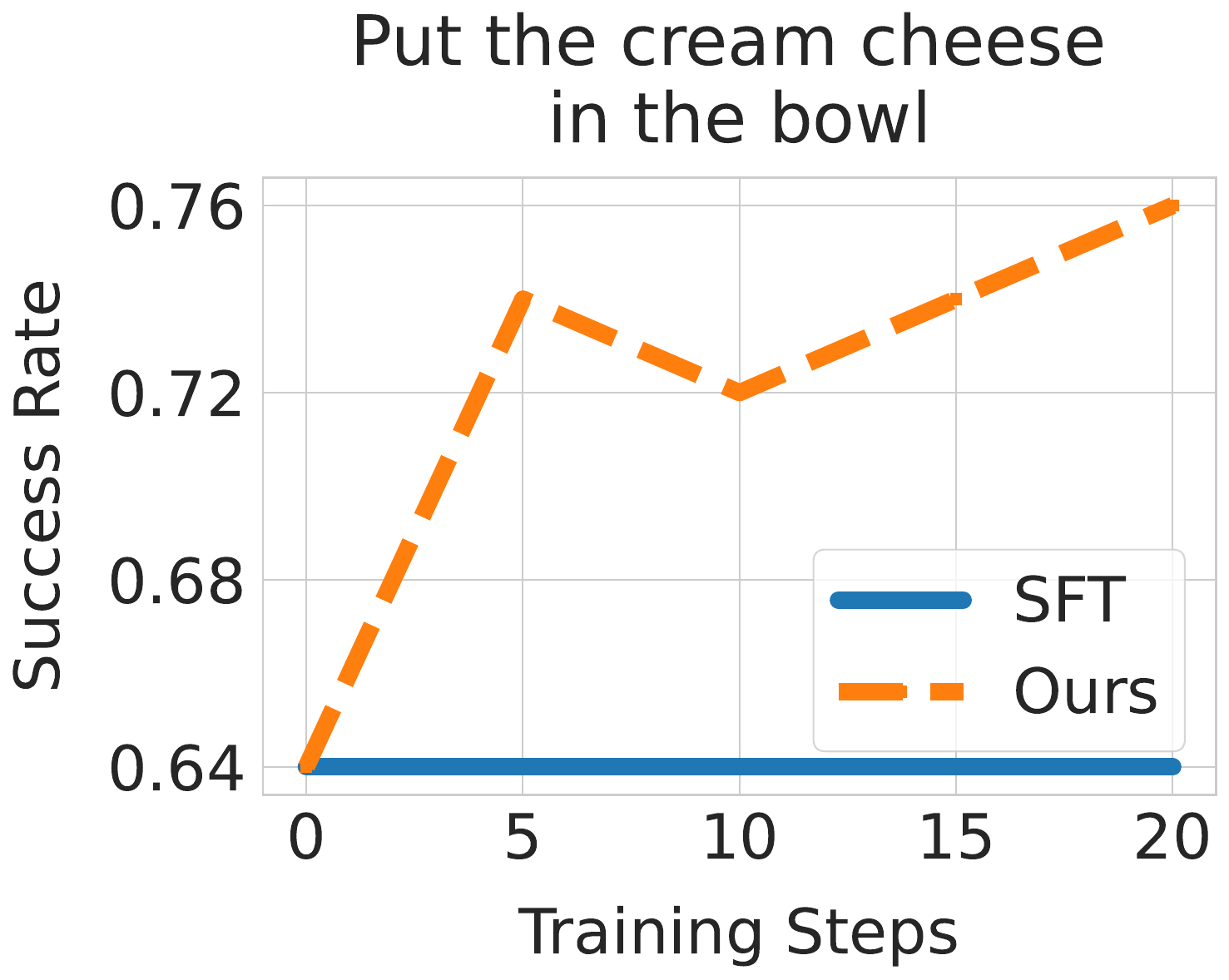}  \\
  \end{minipage}
  \hfill
  \begin{minipage}{0.32\textwidth}
    \centering
    \includegraphics[width=\linewidth]{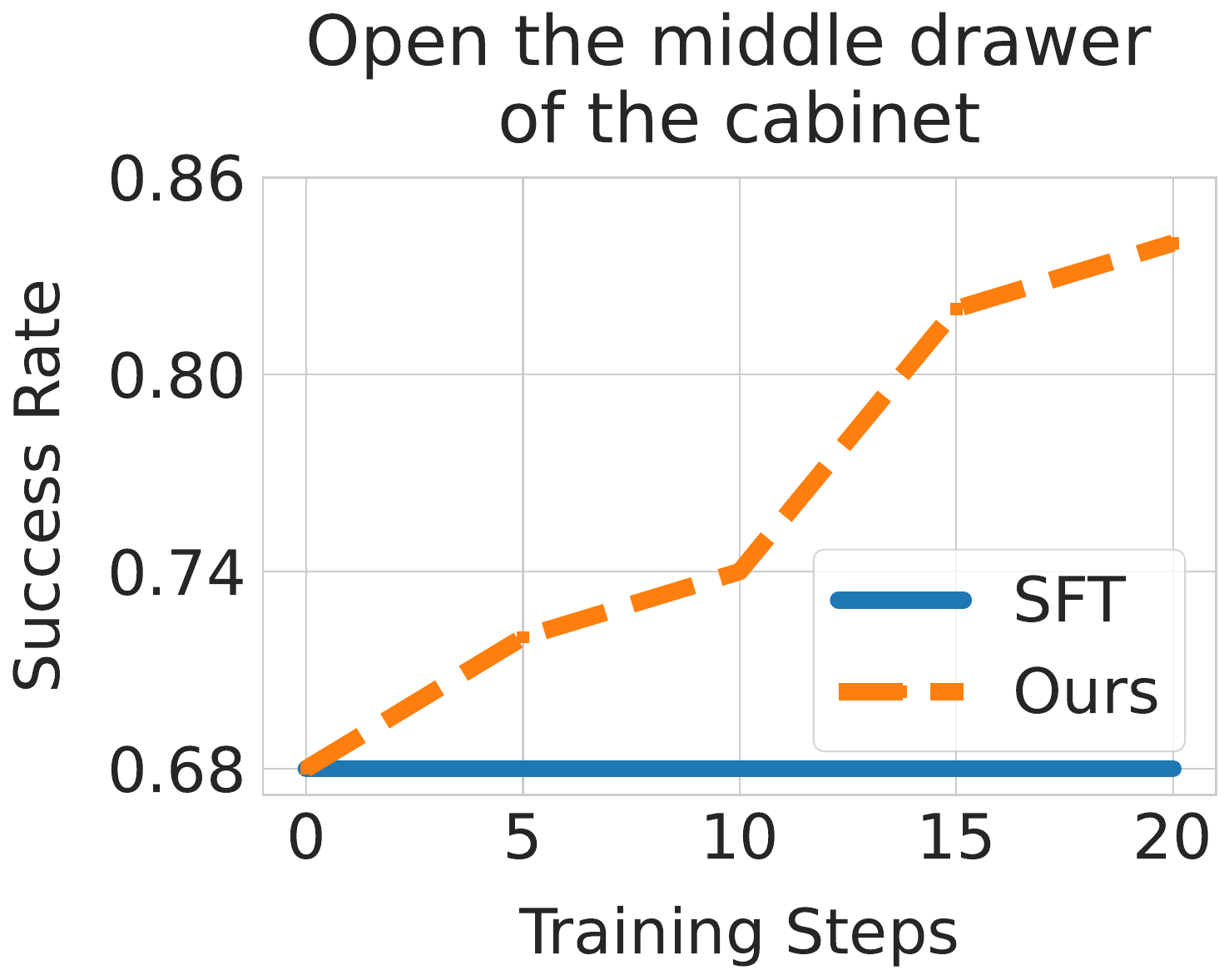}\\
  \end{minipage}
  \hfill
  \begin{minipage}{0.32\textwidth}
    \centering
    \includegraphics[width=\linewidth]{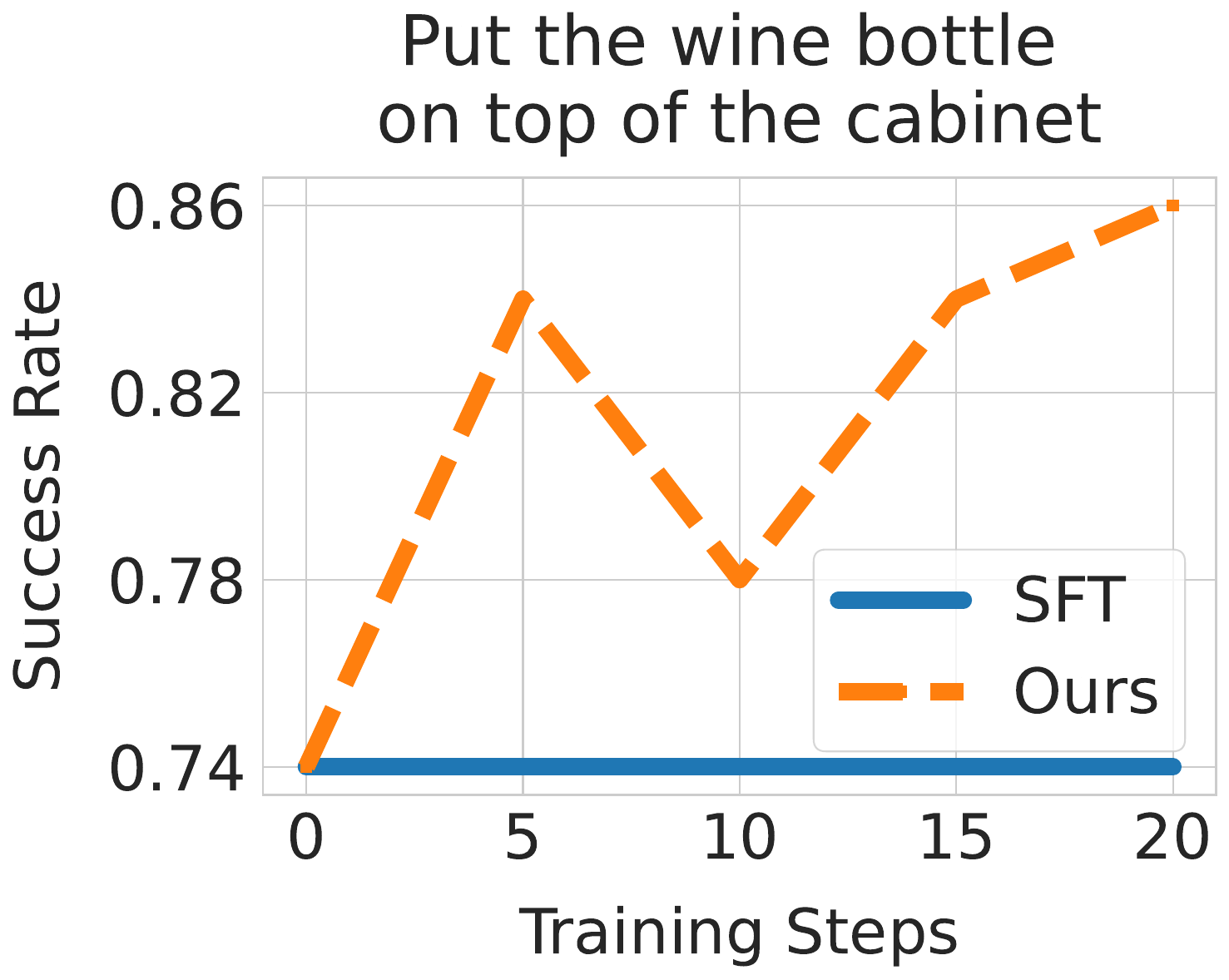}\\
  \end{minipage}
  \vspace{-2mm}
\caption{\textbf{Comparison with SFT on multi-goal tasks.} Note that all results are collected every 5 training steps for three distinct goals.}
\label{fig:line}
\end{figure*}

\subsection{\xiao{Physically-Consistent World Simulator}}
\label{sec:world_model}


\xiao{Figure~\ref{fig:wm_pipeline} presents an overview of our world simulator, which is designed to predict future observations conditioned on a given action sequence. Akin to~\cite{jiang2025enerverseacenvisioningembodiedenvironments}, we first translate the predicted action into a proprioceptive state $\mathbf{s}_{t+1}$ via forward kinematics, and project it onto the image plane to generate an action map. This action map comprises a foreground marker that encodes the projected pose (e.g., position and orientation) and a uniformly black background, which serves to maximize visual contrast and minimize interference with the underlying scene content. The resulting action map is then combined with a historical observation sampled from a memory bank and jointly injected into a U-Net based denoising diffusion network as pixel-level conditioning signals. To ensure that the synthesized future frames remain physically plausible and geometrically coherent with the reference observation, we introduce a geometry-aware feature injection mechanism. Specifically, we extract complementary features from two pre-trained encoders: (i) VGGT~\cite{wang2025vggt}, which excels at preserving fine-grained geometric structures and spatial layouts of the reference image; and (ii) CLIP~\cite{radford2021CLIP}, which captures high-level semantic and contextual information. These features are integrated into the denoising U-Net through cross-attention layers at multiple resolutions, enabling the model to simultaneously respect both local geometric fidelity and global semantic consistency during future frame synthesis. This dual-path injection strategy not only enhances the realism of generated observations but also improves temporal coherence and physical plausibility in long-horizon predictions.}

To train the world model, we find that relying solely on expert demonstrations from the LIBERO benchmark~\cite{liu2023libero} limits generalization to unseen state-action sequences. To address this, we augment the training data by enabling autonomous exploration in the LIBERO simulator. Specifically, we deploy the supervised fine-tuned OpenVLA-OFT policy~\cite{kim2025openvla-oft} to predict actions and execute them in the simulator, which yields the corresponding next proprioceptive state $\mathbf{s}_{t+1}$ and observation $\mathbf{o}_{t+1}$. To further enhance data diversity, we introduce controlled stochasticity by training a scale head that predicts the log-scale parameter $\boldsymbol{\beta}_t$ of a Laplace distribution, with the OpenVLA-OFT action $\boldsymbol{\mu}_t$ as the location parameter: $\mathbf{a}_t \sim \text{Laplace}(\boldsymbol{\mu}_t, \boldsymbol{\beta}_t)$. These perturbed actions are executed to collect additional $(\mathbf{o}_t, \mathbf{s}_t, \mathbf{a}_t, \mathbf{s}_{t+1}, \mathbf{o}_{t+1})$ transition tuples. Finally, we combine these autonomously collected trajectories with the original human-demonstrated successful trajectories from LIBERO~\cite{liu2023libero} to form a diverse and robust training dataset for the world simulator. Additional analysis of the world simulator are provided in the supplementary material.

\subsection{VLM-Guided Instant Reflector}
\label{sec:reward_model}
Previous methods~\cite{tan2025interactiveposttrainingvisionlanguageactionmodels,lu2025vlarlmasterfulgeneralrobotic} rely on simulators to provide binary task success signals, using sparse discrete rewards for RL post-training. These approaches suffer from a key limitation: the lack of termination-aware feedback, causing policies to often continue executing redundant actions after task completion (e.g., over-scooping after object placement). To address this, we propose a VLM-guided instant reflector that leverages a pretrained vision-language model to provide a continuous-valued reward signal.

Given a video of imagined observations $\mathbf{o}_{1:t}$ and a language instruction $\mathbf{g}$, the instant reflector predicts a step-wise reward $R(\mathbf{o}_{1:t}, \mathbf{g}) \in [0,1]$ for each time step $t$, which estimates the probability that the task has been successfully completed by time $t$. The architecture consists of a frozen vision encoder $\mathcal{E}_{\text{vision}}$ that extracts patch embeddings from video frames, a frozen LLM $\mathcal{E}_{\text{LLM}}$ that performs cross-modal reasoning over the visual-language sequence, and a lightweight reward head $\mathcal{R}_\theta$ that computes:
\begin{equation}
    R(\mathbf{o}_{1:t}, \mathbf{g}) = \sigma(\mathcal{R}_\theta(h_t)),
\label{eq:reward}
\end{equation}
where $\sigma$ is the sigmoid function and $h_t$ is the pooled multimodal embedding from the LLM at time $t$. The termination signal is triggered at the timestep $t$ where $R(\mathbf{o}_{1:t}, \mathbf{g}) > \eta$, with threshold $\eta = 0.5$.

For training, we utilize per-frame binary success labels: for each trajectory, every timestep $t$ is annotated with $y_t \in \{0,1\}$, indicating whether the task is completed at or before $t$. These labels are derived from two sources: (1) expert trajectories from the LIBERO dataset~\cite{liu2023libero}, where success is determined by task-specific criteria, and (2) policy-generated trajectories collected in simulator (Section~\ref{sec:world_model}). The reward head $\mathcal{R}_\theta$ is trained with BCE loss:
\[
\mathcal{L} = \text{BCE}\big(R(\mathbf{o}_{1:t}, \mathbf{g}), y_t\big).
\]
This supervision enables the reflector to recognize task completion as soon as it occurs. During RL, we use the reward sparsely: the return is computed using a single reward assigned at the termination timestep (or at $T$ if no termination occurs).

\xiao{This design enables \model{} to support real-time termination while achieving both high sample efficiency and stable policy learning. In contrast, prior VLA post-training approaches~\cite{tan2025interactiveposttrainingvisionlanguageactionmodels,lu2025vlarlmasterfulgeneralrobotic} rely on binary rewards (e.g., 1 for success, 0 for failure), which lead to degenerate advantage estimates when rollout trajectories are homogeneous, i.e., when all episodes succeed or all fail. In such cases, the empirical advantage values collapse to zero across the batch, providing no learning signal for policy updates and drastically reducing training efficiency. By instead employing a continuous reward signal in the range $[0, 1]$ that reflects fine-grained task progress, our method ensures non-trivial advantage estimates. This eliminates the need to balance successful and failed rollouts during data collection, enabling more efficient use of samples.}
\begin{figure*}[t]
\begin{center}
\includegraphics[width=1\textwidth]{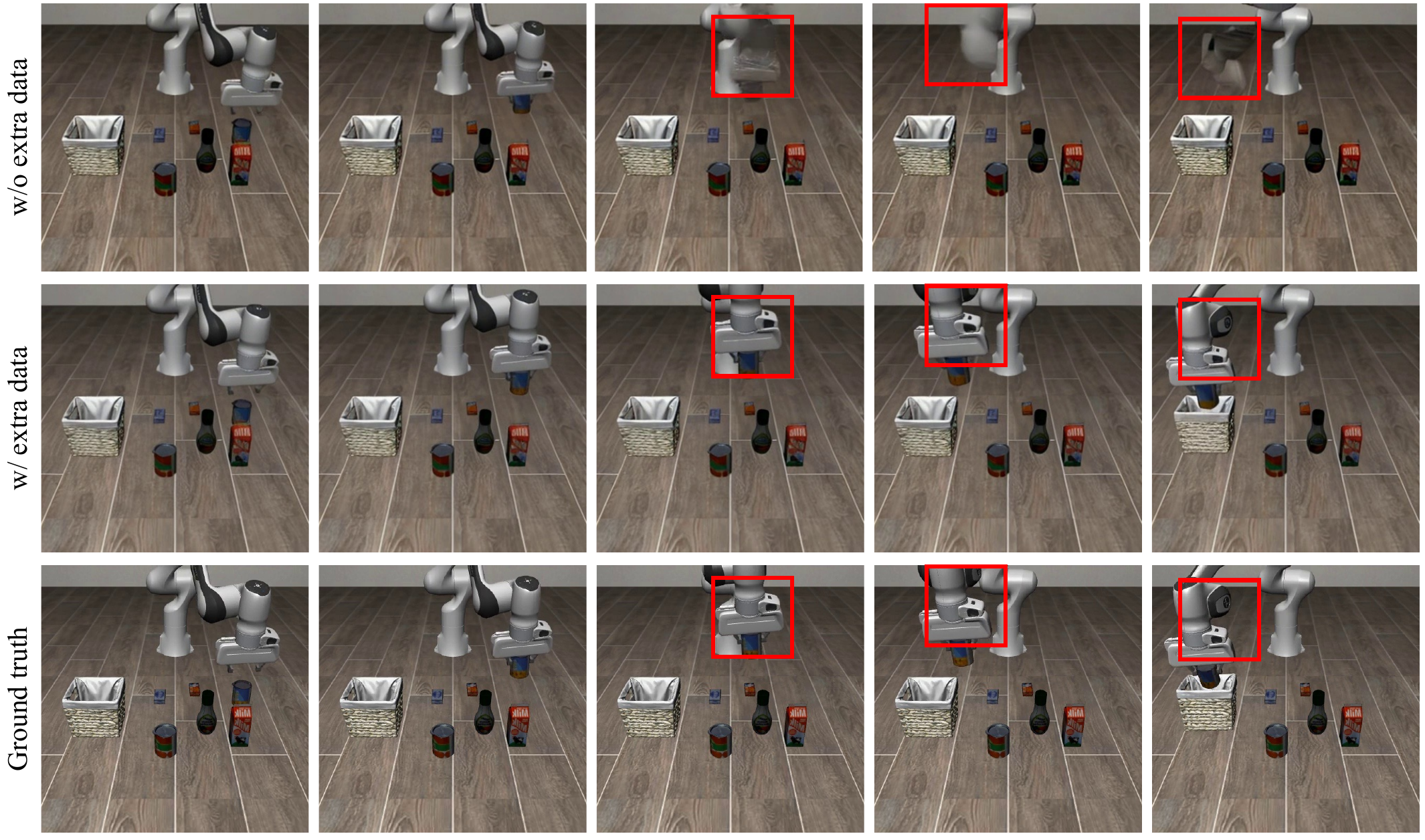} \\
\end{center}
 \vspace{-4mm}
\caption{\textbf{Rendering comparison of world simulator trained with and without extra data.}}
\label{fig:wm}
\end{figure*}

\subsection{Post Training of VLA Model}
Our reinforcement learning pipeline employs a RLOO~\cite{ahmadian2024back} objective with continuous reward signals; the full algorithm is provided in the supplementary material. Following~\cite{tan2025interactiveposttrainingvisionlanguageactionmodels}, training proceeds in three stages: rollout generation, advantage estimation, and policy optimization.

During rollout, we generate trajectories $\tau = (\mathbf{o}_{1:T}, \mathbf{s}_{1:T}, \mathbf{g}, \mathbf{a}_{1:T})$ using the world simulator (Section~\ref{sec:world_model}). Starting from an initial observation $\mathbf{o}_1$, proprioceptive state $\mathbf{s}_1$, and language instruction $\mathbf{g}$, the deterministic VLA policy $\pi_\theta$ predicts a base action $\boldsymbol{\mu}_t = \pi_\theta(\mathbf{o}_{1:t}, \mathbf{s}_{1:t}, \mathbf{g})$. A separate scale head, trained to model action uncertainty, outputs a log-scale parameter $\boldsymbol{\beta}_t$. Together, they define a factorized Laplace distribution, from which the executed action $\mathbf{a}_t$ is sampled. This enables adaptive, uncertainty-aware exploration. The world simulator then predicts the next observation $\mathbf{o}_{t+1}$ using proprioceptive state $\mathbf{s}_{t+1}$. The VLM-guided instant reflector evaluates the partial visual trajectory $\mathbf{o}_{1:t+1}$ and outputs a step-wise reward $R(\mathbf{o}_{1:t+1}, \mathbf{g}) \in [0,1]$. Rollout terminates either at the maximum timestep $T$ or when $R(\mathbf{o}_{1:t+1}, \mathbf{g}) > \eta$. For RL, we assign a single trajectory-wise reward $R_n = R(\mathbf{o}_{1:t_\text{end}}, \mathbf{g})$, where $t_\text{end}$ is the termination or final timestep.

We adopt Leave-One-Out Proximal Policy Optimization (LOOP)~\cite{chen2025reinforcement} that combines RLOO~\cite{ahmadian2024back} based advantage estimation and PPO~\cite{schulman2017proximal} for policy updating. For each initial state, we generate $N$ rollouts $\{\tau_1, \dots, \tau_N\}$ using a fixed behavior policy $\pi_\phi$ (the policy at the beginning iteration). Each trajectory receives a scalar reward $R_n$ from the instant reflector. The RLOO baseline for trajectory $n$ is the average reward of the other rollouts:
\begin{equation}  
b_n = \frac{1}{N-1} \sum_{j \neq n} R_j, \quad A_n = R_n - b_n,
\end{equation}  
where $A_n$ is the trajectory-wise advantage. To update the policy, we treat both the current and behavior policies as inducing stochastic action distributions via their action and scale heads. The importance ratio at timestep $t$ of trajectory $n$ is computed as:
\[
r_{t,n} = \frac{p_\theta(\mathbf{a}_{t,n} \mid \mathbf{o}_{t,n}, \mathbf{s}_{t,n}, \mathbf{g}_n)}{p_\phi(\mathbf{a}_{t,n} \mid \mathbf{o}_{t,n}, \mathbf{s}_{t,n}, \mathbf{g}_n)},
\]
where $p_\theta$ and $p_\phi$ denote the action distributions induced by the current policy $\pi_\theta$ and behavior policy $\pi_\phi$, respectively, each modeled as a product of independent Laplace distributions over action dimensions. The policy is optimized via the clipped PPO objective:
\begin{equation}\label{eq:ppo}
\mathcal{L}_{\text{PPO}} =  -\min(r_{t,n} A_n, \, \text{clip}(r_{t,n}, 1-\epsilon, 1+\epsilon) A_n ) ,
\end{equation}  
with $\epsilon$ representing to the clipping threshold. Note that the advantage $A_n$ is broadcasted to all timesteps.

\begin{table}[t]
\centering
\caption{\textbf{Comparison with simulator-based RL method.}}
\vspace{-2mm}
\resizebox{0.9\linewidth}{!}
{
\begin{tabular}{l c c c c}
\toprule[1pt]
 \multicolumn{1}{l}{Method}& \multicolumn{1}{c}{Goal}& \multicolumn{1}{c}{Object}& \multicolumn{1}{c}{Spatial}& \multicolumn{1}{c}{Long}\\ 
\midrule
RIPT-VLA~\cite{tan2025interactiveposttrainingvisionlanguageactionmodels}&86.2&83.4&88.6&58.4\\
Ours&86.4&86.6&87.6&57.8\\
\bottomrule[1pt]
\end{tabular}
}

\label{table:com_rl}
\end{table}


\begin{figure*}[h]
  \centering
  \includegraphics[width=0.19\linewidth]{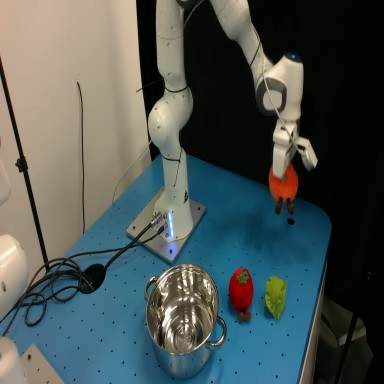}%
  \hfill
  \includegraphics[width=0.19\linewidth]{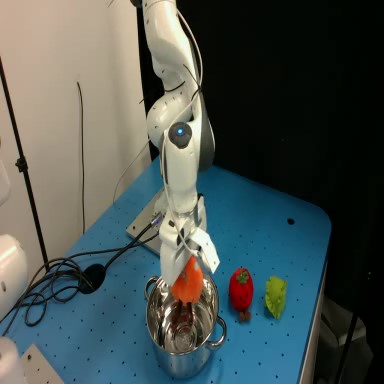}%
  \hfill
  \includegraphics[width=0.19\linewidth]{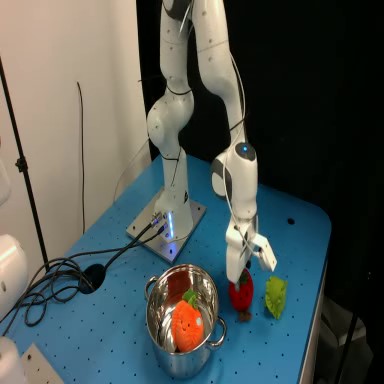}%
  \hfill
  \includegraphics[width=0.19\linewidth]{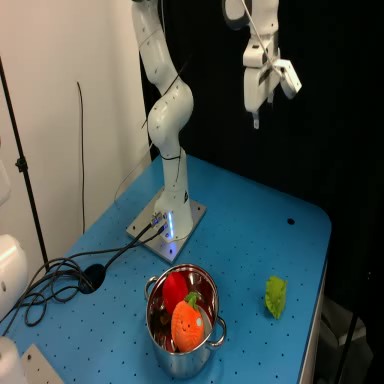}%
  \hfill
  \includegraphics[width=0.19\linewidth]{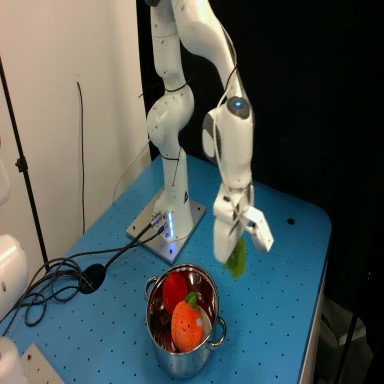}%
  \vspace{-2.0mm}
\caption{\textbf{Real-world experiment on task ``clean table''. Full video result can be found in the supplementary material. }}
\label{fig:real}
\end{figure*}

\subsection{Implementation Details}
Experiments were performed on 8 NVIDIA H20 GPUs (96 GB VRAM each), with a total training time of approximately 48 hours. We adopt LoRA~\cite{hu2022lora} with rank 32 for parameter-efficient fine-tuning of the vision-language backbone, while the action head and scale head are trained with full parameters. We use a batch size of 4. The LoRA adapters are optimized with a learning rate of $1 \times 10^{-4}$, and the action/scale heads are trained with a learning rate of $1 \times 10^{-5}$. We set the number of rollouts per iteration to $N=8$ and the PPO clipping threshold to $\epsilon=0.1$.

\begin{table}[t]
\centering
\resizebox{\linewidth}{!}
{
\begin{tabular}{l c c c c}
\toprule[1pt]
\multirow{1}{*}{Method}& \multicolumn{1}{c}{Clean table}& \multicolumn{1}{c}{Put green toy}& \multicolumn{1}{c}{Put red toy}& \multicolumn{1}{c}{Put orange toy}\\ 
\midrule
OpenVLA-OFT &20 &30&30 &20\\
Ours& 30&50&40 &50\\
\bottomrule[1pt]
\end{tabular}
}

\caption{\textbf{Comparison results of real-world experiments.}}
\vspace{-2.0mm}
\label{table:real}
\end{table}

\section{Experiments}

\vspace{0.5em}
\noindent \textbf{Benchmarks.}
We evaluate our model on the LIBERO benchmark \cite{liu2023libero}, a simulation-based robotic learning platform designed for vision-language manipulation tasks. The benchmark includes four task suites targeting distinct cognitive challenges: LIBERO-Spatial focusing on spatial reasoning via object arrangement; LIBERO-Goal assessing goal-conditioned planning with end-state requirements; LIBERO-Object testing object-centric manipulation across categories; LIBERO-10 (LIBERO-Long) addressing prolonged sequential decision-making. Each suite contains 10 tasks with 50 trajectories for training and 50 for testing per task; we train OpenVLA-OFT using only 5 trajectories from the training split to validate performance under extreme data scarcity, while evaluating on the full trajectory test split to demonstrate the generalization capability.

\vspace{0.5em}
\noindent \textbf{Baselines.}
We compare our method with five state-of-the-art supervised fine-tuning (SFT) methods including $\pi_0$ \cite{black2024pi0visionlanguageactionflowmodel}, 
$\pi_0$ + FAST \cite{pertsch2025fast}, OpenVLA \cite{kim24openvla}, UniVLA \cite{bu2025univla}, OpenVLA-OFT \cite{kim2025openvla-oft} \xiao{and one simulator-based RL method RIPT-VLA~\cite{tan2025interactiveposttrainingvisionlanguageactionmodels}}. For fair evaluation, all baselines are retrained under identical 5-trajectory per-task constraints, with performance metrics reported on the complete test set. 

\subsection{Comparison with State-of-the-art Methods}
Table~\ref{table:main_compare} presents success rate comparison between our method and the baseline models. As shown, our method gains higher task success rate, demonstrating the effectiveness of our proposed post-training strategy. Figure~\ref{fig:line} further compares our method and the supervised fine-tuning (SFT) baseline on multi-goal tasks, where we can see that our approach achieves superior performance within only 20 training steps, clearly outperforming the compared SFT model. This rapid convergence and early dominance highlight the efficiency and effectiveness of our method in learning conditioned policies with minimal training iterations. \xiao{Table~\ref{table:com_rl} presents a comparison with RIPT-VLA~\cite{tan2025interactiveposttrainingvisionlanguageactionmodels}, a simulator-based reinforcement learning approach. Our method achieves comparable performance while offering a crucial practical advantage: it is readily deployable in real-world settings. In contrast, RIPT-VLA is limited to simulated environments.}

\begin{figure*}[t]
\centering
    \begin{subfigure}[b]{0.19\textwidth}  
        \centering
        \includegraphics[width=\textwidth]{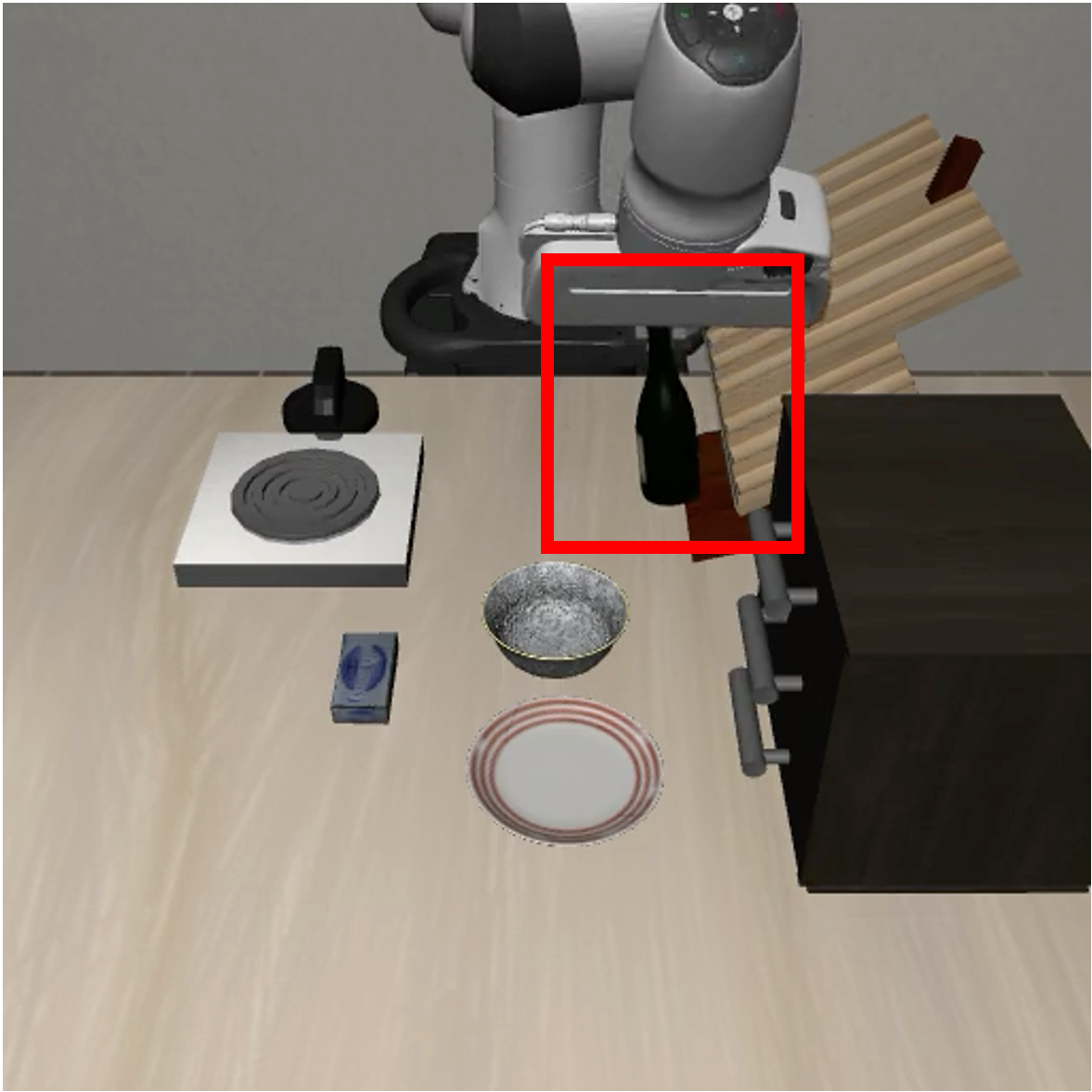}  
        \caption{Executing}
    \end{subfigure}
    \hfill  
    \begin{subfigure}[b]{0.19\textwidth}
        \centering
        \includegraphics[width=\textwidth]{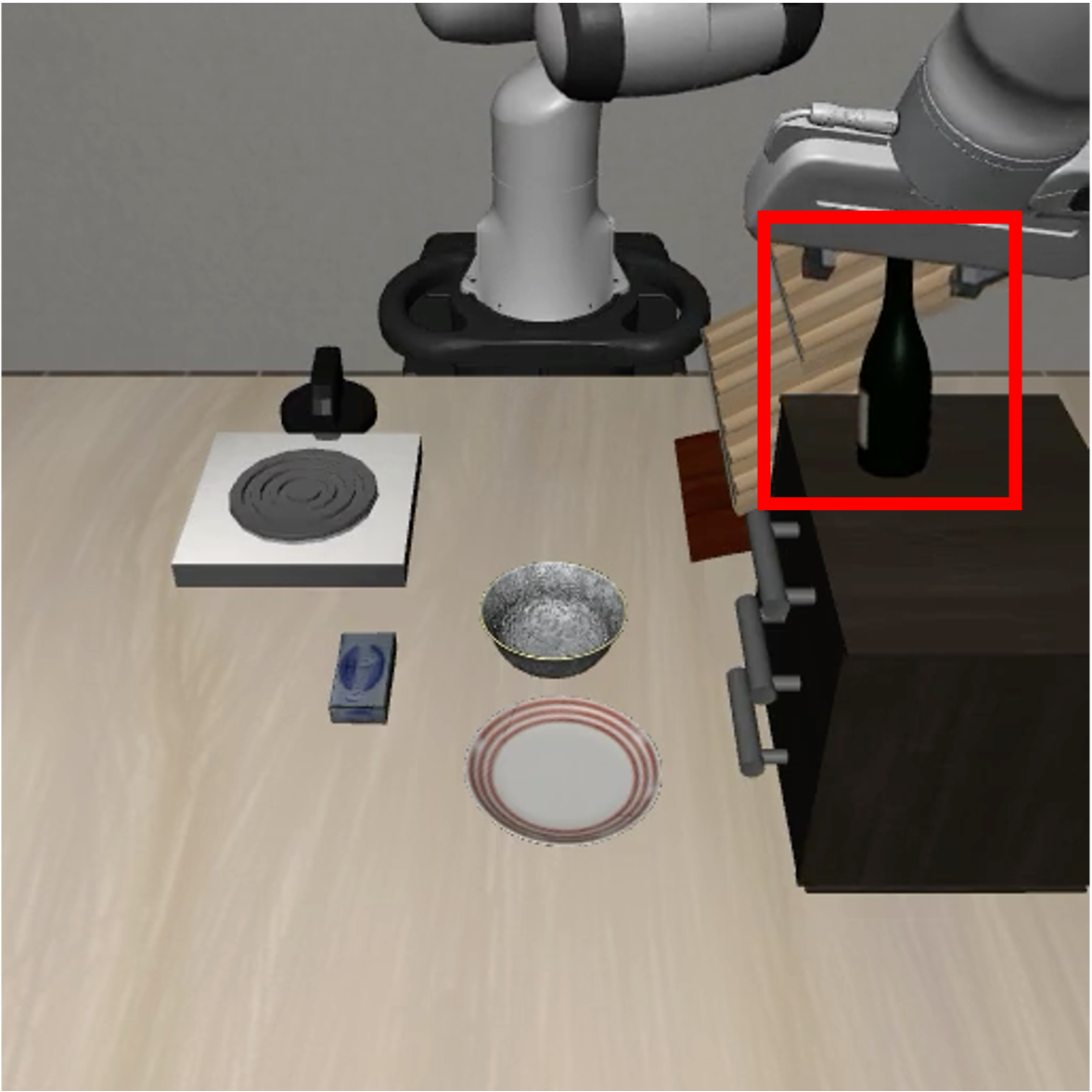}
        \caption{Success}
    \end{subfigure}
    \hfill  
    \begin{subfigure}[b]{0.19\textwidth}
        \centering
        \includegraphics[width=\textwidth]{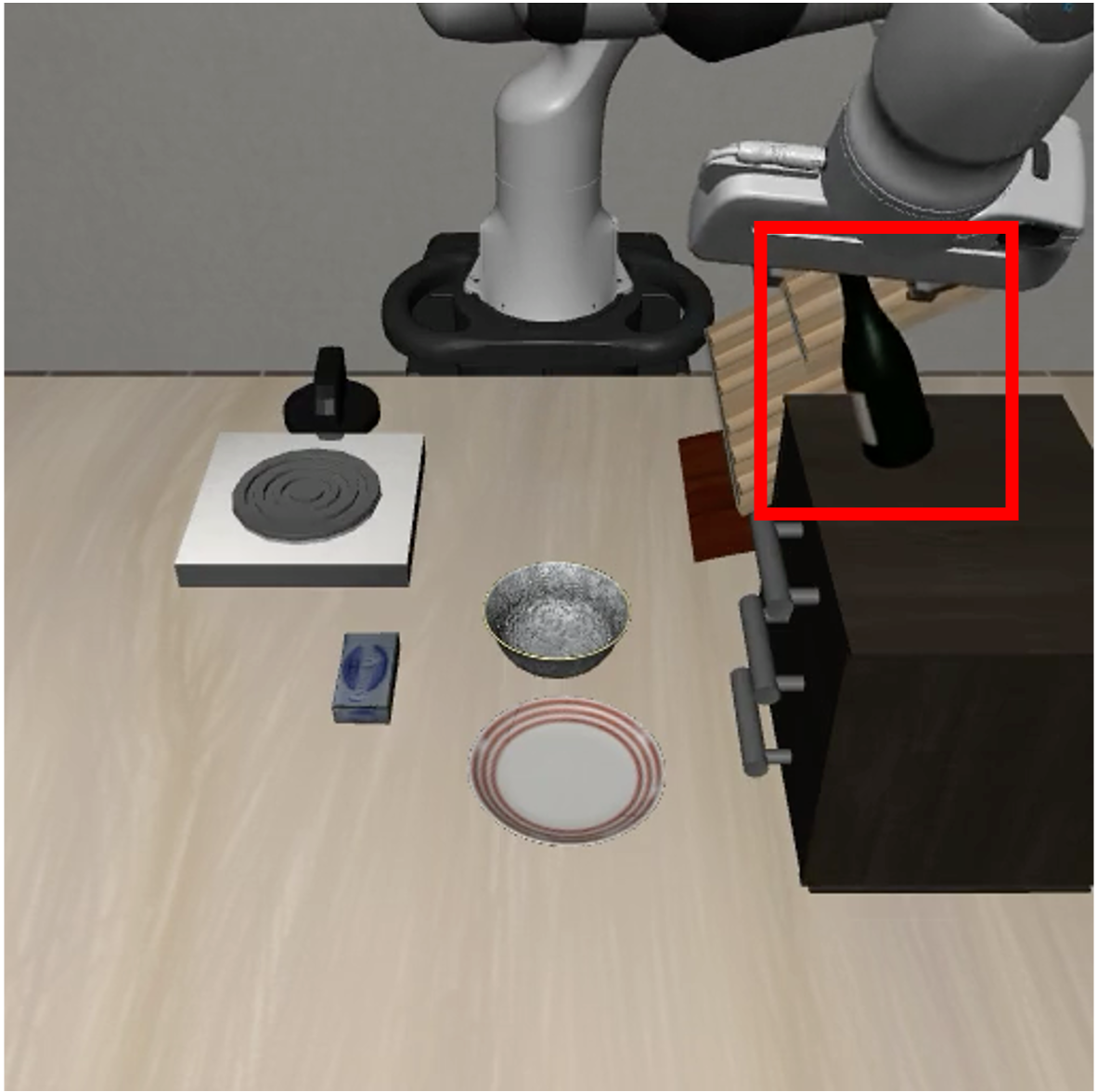}
        \caption{Fail}
    \end{subfigure}
    \hfill  
    \begin{subfigure}[b]{0.19\textwidth}
        \centering
        \includegraphics[width=\textwidth]{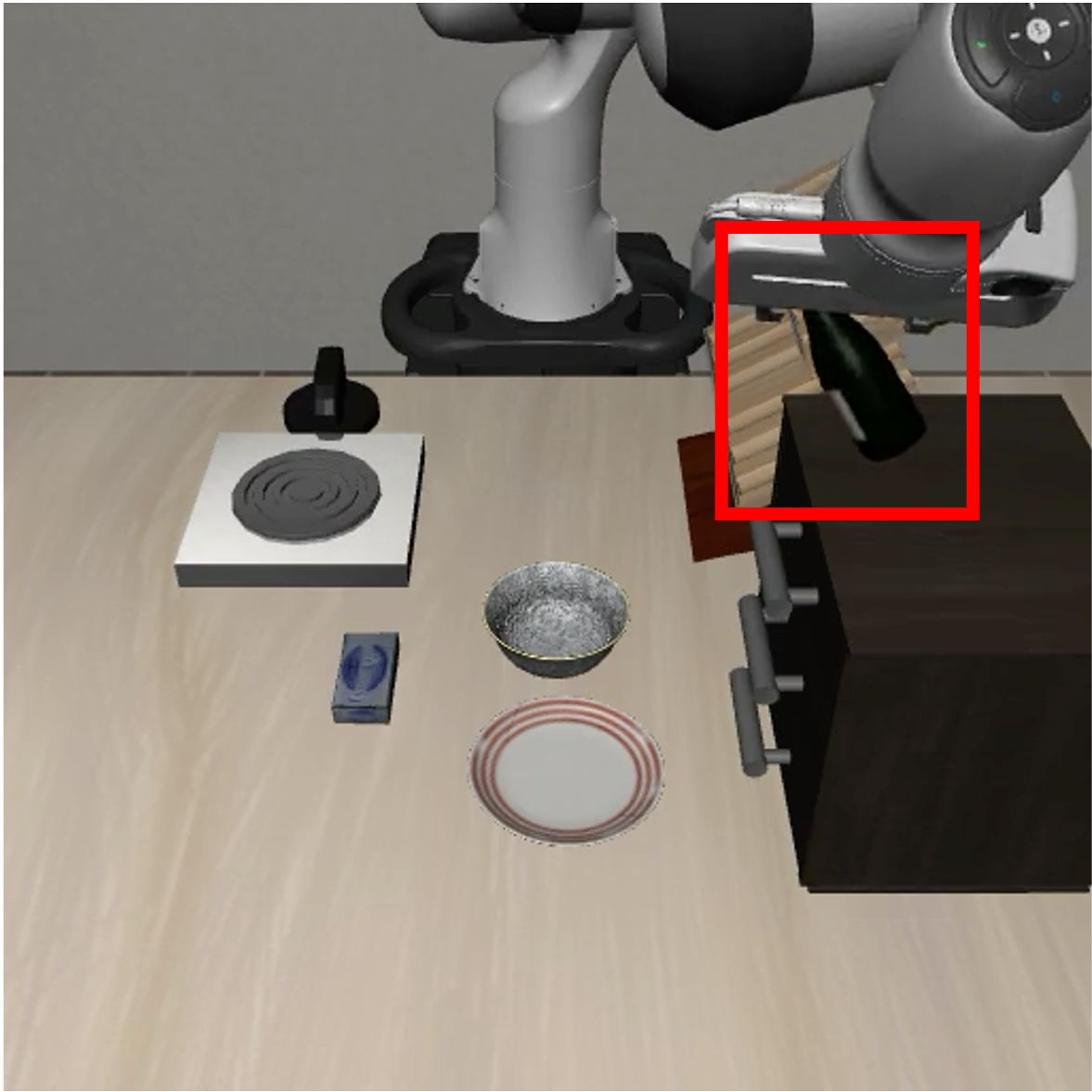}
        \caption{Fail}
    \end{subfigure}
    \hfill  
    \begin{subfigure}[b]{0.19\textwidth}
        \centering
        \includegraphics[width=\textwidth]{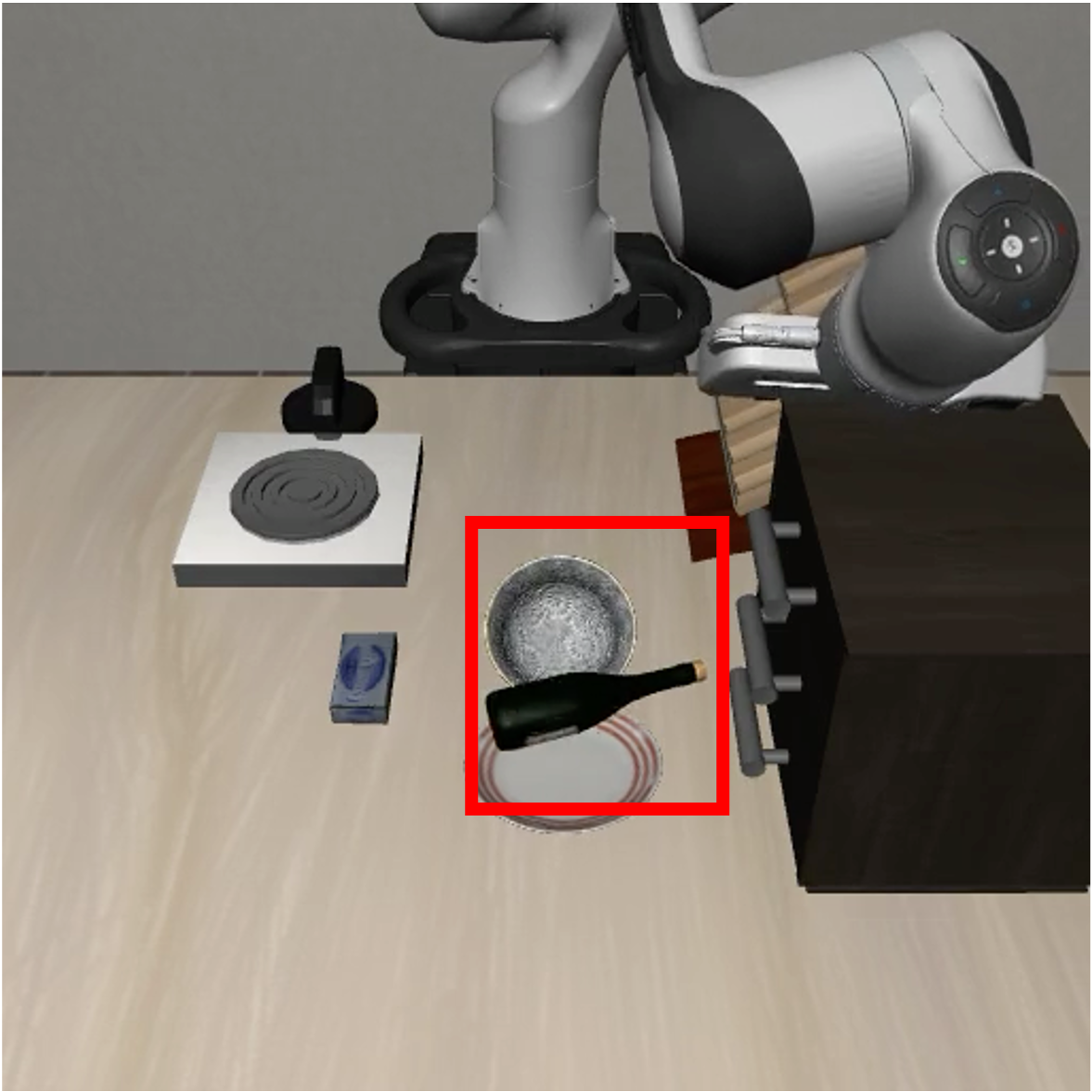}
        \caption{Fail}
    \end{subfigure}

\vspace{-1mm}
\caption{\textbf{Post-success failure in VLA execution.} An illustrative example for ``put the wine bottle on top of the cabinet'' shows the VLA model completes the task (a-b), but fails due to delayed termination (c-e), validating the necessity of dynamic termination mechanism.}
\label{fig:post_success}
\end{figure*}

\begin{table*}[!t]
\centering
\caption{\textbf{Comparison of task termination strategy under realistic feedback constraints.} Note, all compared methods are evaluated under the setting where ground-truth termination feedback is unavailable, while our method autonomously detects task completion via the proposed reward model. Success rates are measured when reaching the maximum action steps.}
\resizebox{\linewidth}{!}
{
\begin{tabular}{l c c c c c}
\toprule[1pt]
\multirow{1}{*}{Method}& \multicolumn{1}{c}{LIBERO-Goal}& \multicolumn{1}{c}{LIBERO-Object}& \multicolumn{1}{c}{ LIBERO-Spatial}&\multicolumn{1}{c}{LIBERO-Long}&\multicolumn{1}{c}{Average}\\ 
\midrule
$\pi_0$~\cite{black2024pi0visionlanguageactionflowmodel}&55.4 & 71.0&72.6 & 20.6& 54.9 \\
$\pi_0$+FAST~\cite{pertsch2025fast}&21.2 & 74.0& 44.8& 15.0& 38.75 \\
OpenVLA~\cite{kim24openvla}&68.4  & 47.4 & 59.8 & 26.6 &50.55 \\
UniVLA~\cite{bu2025univla}& 72.0 & 75.2 &  66.4& 48.0 &  65.4 \\
OpenVLA-OFT~\cite{kim2025openvla-oft}& 67.4 & 73.8 & 71.2 & 39.8 & 63.05  \\
OpenVLA-OFT + Post-training (Ours)&\textbf{85.0} & \textbf{78.4}& \textbf{78.4}&\textbf{57.8} & \textbf{74.9} \\
\bottomrule[1pt]
\end{tabular}
}
\vspace{-2mm}
\label{table:termi_compare}
\end{table*}


\begin{table}[t]
\centering
\caption{\textbf{Ablation studies.} We evaluate how the extra training data for world simulator learning and the reward head for trajectory scoring affect the performance of our method.}
\vspace{-2mm}
\resizebox{\linewidth}{!}
{
\begin{tabular}{c c c c c c}
\toprule[1pt]
 \multicolumn{1}{c}{Extra Data}& \multicolumn{1}{c}{Reward Head}& \multicolumn{1}{c}{Goal}& \multicolumn{1}{c}{Object}& \multicolumn{1}{c}{Spatial}& \multicolumn{1}{c}{Long}\\ 
\midrule
 & & 68.4& 75.2& 73.2&42.2 \\
 \checkmark & &79.8 & 81.8& 78.4& 44.6\\
 &\checkmark& 68.8&76.4&74.4&43.8 \\
\checkmark & \checkmark& \textbf{86.4} &\textbf{86.6} &\textbf{87.6} & \textbf{57.8} \\
\bottomrule[1pt]
\end{tabular}
}

\label{table:ablation}
\end{table}


\subsection{Real-world Experiment}
We conduct real-world experiments across four tasks, \ie, ``clean table'', ``put green toy in cabinet'', ``put red toy in cabinet'', and ``put orange toy in cabinet'', Figure~\ref{fig:real} shows several sampled frames from the ``clean table'' task, where the goal is to pick up three toys and place them into the bucket. For each task, we collect 10 trajectories for both policy SFT and world model fine-tuning. As shown in the table \ref{table:real}, our method outperforms OpenVLA-OFT on all four tasks, validating that our framework can be effectively transferred to real-world environments.

\subsection{Ablation Studies}
\vspace{0.5em}
\noindent \textbf{Effect of World Simulator.}
We study how the world simulator’s generative quality affects performance. Figure~\ref{fig:wm} compares two variants: (1) w/o extra: trained only on successful human trajectories; and (2) w/ extra: augmented with our data containing both success and failure cases. Without exposure to suboptimal actions, the former fails to model resulting object states, leading to poor tracking under VLA prediction errors. Our enhanced simulator substantially improves arm tracking and interaction fidelity. Table~\ref{table:ablation} confirms that low-fidelity simulation degrades VLA training.  Additional analysis, ablations, and videos are in the supplementary material.

\vspace{0.5em}
\noindent \textbf{Effect of Instant Reflector.} 
Table~\ref{table:ablation} compares two variants: (1) w/o reward head: using a pre-trained VLM with prompt-based binary classification; and (2) w/ reward head: a trainable module scoring actions on a continuous scale. Experiments show that off-the-shelf VLMs offer limited gains and can impair learning in complex tasks due to misalignment with fine-grained action evaluation. In contrast, our reward head provides more accurate and reliable action assessment.

\vspace{0.5em}
\noindent \textbf{Effect of Termination Signals.}
Table~\ref{table:termi_compare} further validates our method’s ability to detect task success. Unlike conventional approaches, which rely on provided termination signals, our framework uses a VLM-guided instant reflector to dynamically assess completion and stop execution promptly. To isolate this advantage, we force all baselines to adhere strictly to the full horizon, while our method terminates based on reflector predictions. As shown, baselines suffer performance degradation due to redundant post-success actions that perturb object states (see Figure~\ref{fig:post_success}). In contrast, our approach preserves task outcomes by halting immediately upon success detection, demonstrating the reflector’s effectiveness in enabling timely termination.

\subsection{Limitations and Future Work}
Despite the effectiveness of our approach in enhancing VLA manipulation capabilities, several limitations remain. First, both the world simulator and the instant reflector rely on diverse training data to achieve high-fidelity simulation and accurate task evaluation. We anticipate that future advances in general-purpose world models will alleviate this dependency. Second, policy optimization in our framework is currently slower than in concurrent methods due to computational bottlenecks in simulator-based trajectory generation. Addressing these challenges through more efficient simulation will be a key focus of our future work.

\section{Conclusion}
We present \model{}, a post-training framework for Vision-Language-Action (VLA) models that eliminates the need for costly and potentially unsafe interactions with physical environments. By leveraging a physically consistent world simulator, enhanced with a geometry-aware feature injection strategy, \model{} enables safe and low-cost exploration in virtual settings. Furthermore, our VLM-guided instant reflector supports dynamic task-aware termination, preventing redundant post-success actions that could otherwise disrupt task outcomes. Extensive experiments on complex robotic manipulation tasks demonstrate that \model{} achieves superior performance, particularly in low-data regimes.

\maketitlesupplementary
\section{Algorithm}
\begin{algorithm*}[t]
\caption{\model{} Training Algorithm}
\label{alg:main}
\begin{algorithmic}[1]
\Input Pretrained VLA policy $\pi_\theta$, scale head $\beta_\theta$, VLM-based reward function $R(\mathbf{o}_{1:t}, \mathbf{g})$, context dataset $\mathcal{D}_{\text{context}}$
\For{training iteration $= 1$ to $M$}
    \State Set behavior policy: $\pi_\phi \gets \pi_\theta$, \quad $\beta_\phi \gets \beta_\theta$  \Comment{Fix old policy and scale head}
    \State Initialize rollout buffer $\mathcal{D}_{\text{rollout}} \gets \emptyset$
    \While{$|\mathcal{D}_{\text{rollout}}| < B$} \Comment{Rollout Collection}
        \State Sample context $\mathbf{c} = (\mathbf{g}, \mathbf{o}_1, \mathbf{s}_1) \sim \mathcal{D}_{\text{context}}$
        \For{$n = 1$ to $N$} \Comment{Generate $N$ rollouts per context}
            \State Initialize trajectory $\tau_n \gets (\mathbf{o}_1, \mathbf{s}_1)$
            \For{$t = 1$ to $T$}
                \State Predict base action: $\boldsymbol{\mu}_t \gets \pi_\phi(\mathbf{o}_{1:t}, \mathbf{s}_{1:t}, \mathbf{g})$
                \State Predict log-scale: $\boldsymbol{\beta}_t \gets \beta_\phi(\mathbf{o}_{1:t}, \mathbf{s}_{1:t}, \mathbf{g})$
                \State Sample action: $\mathbf{a}_t \sim \text{Laplace}(\boldsymbol{\mu}_t, \exp(\boldsymbol{\beta}_t))$
                \State Compute next proprioceptive state: $\mathbf{s}_{t+1} \gets \text{FK}(\mathbf{s}_t, \mathbf{a}_t)$
                \State Predict next observation: $\mathbf{o}_{t+1} \gets \text{WorldSim}(\mathbf{o}_t, \mathbf{s}_{t+1})$
                \State Append $(\mathbf{a}_t, \mathbf{o}_{t+1}, \mathbf{s}_{t+1})$ to $\tau_n$
                \If{$R(\mathbf{o}_{1:t+1}, \mathbf{g}) > \eta$} \Comment{Termination check ($\eta=0.5$)}
                    \State $t_{\text{end}} \gets t+1$; \textbf{break}
                \EndIf
            \EndFor
            \State Set trajectory reward: $R_n \gets R(\mathbf{o}_{1:t_{\text{end}}}, \mathbf{g})$
            \State Store log-probabilities $\log p_\phi(\mathbf{a}_{1:t_{\text{end}}} \mid \cdot)$ for importance weighting
        \EndFor
        \State Compute RLOO baselines: $b_n \gets \frac{1}{N-1} \sum_{j \neq n} R_j$ for all $n$
        \State Compute advantages: $A_n \gets R_n - b_n$
        \State Add $\{(\tau_n, A_n, \log p_\phi(\cdot))\}_{n=1}^N$ to $\mathcal{D}_{\text{rollout}}$
    \EndWhile
    \For{optimization step $= 1$ to $K$}
        \State Sample batch from $\mathcal{D}_{\text{rollout}}$
        \State Compute current log-probabilities $\log p_\theta(\mathbf{a} \mid \cdot)$
        \State Compute importance ratios: $r_t \gets \exp(\log p_\theta - \log p_\phi)$
        \State Update $\pi_\theta$ and $\beta_\theta$ by minimizing PPO loss
    \EndFor
\EndFor
\end{algorithmic}
\end{algorithm*}

We show the full reinforcement learning post-training algorithm in Algorithm \ref{alg:main}.

\section{Comparison to Octo.}
We compare with Octo~\cite{octomodelteam2024octoopensourcegeneralistrobot} in Table~\ref{table:octo}. As shown in the table, our method outperforms Octo across all four LIBERO suites despite using only $10\%$ training data, manifesting its effectiveness. 
\begin{table}[h]
\centering
\resizebox{\linewidth}{!}
{
\begin{tabular}{l c c c c}
\toprule[1pt]
\multirow{1}{*}{Method}& \multicolumn{1}{c}{LIBERO-Goal}& \multicolumn{1}{c}{LIBERO-Object}& \multicolumn{1}{c}{ LIBERO-Spatial}& \multicolumn{1}{c}{ LIBERO-Long}\\ 
\midrule
Octo &84.6 &85.7&78.9&51.1\\
Ours&\textbf{86.4} &\textbf{86.6}&\textbf{87.6}&\textbf{57.8}\\
\bottomrule[1pt]
\end{tabular}
}
\caption{\textbf{Quantitative comparison with Octo on LIBERO.}}
\label{table:octo}
\end{table}

\section{More Implementation Details}
\label{sec:impl}
\subsection{Deatils of Scale Head}
Our method builds upon OpenVLA-OFT~\citep{kim2025openvla-oft}, which predicts continuous actions via an action head that takes hidden states $f \in \mathbb{R}^{d}$ as input and employs L1 loss for action regression:
\begin{equation}
    \mathcal{L}_{\text{L1}} = \|\mathbf{a}_{\text{gt}} - \boldsymbol{\mu}\|_1 \quad \text{where } \boldsymbol{\mu} = \text{MLP}_{\text{action}}(f).
\end{equation}

\begin{figure}[t]
  \centering
  \includegraphics[width=\linewidth]{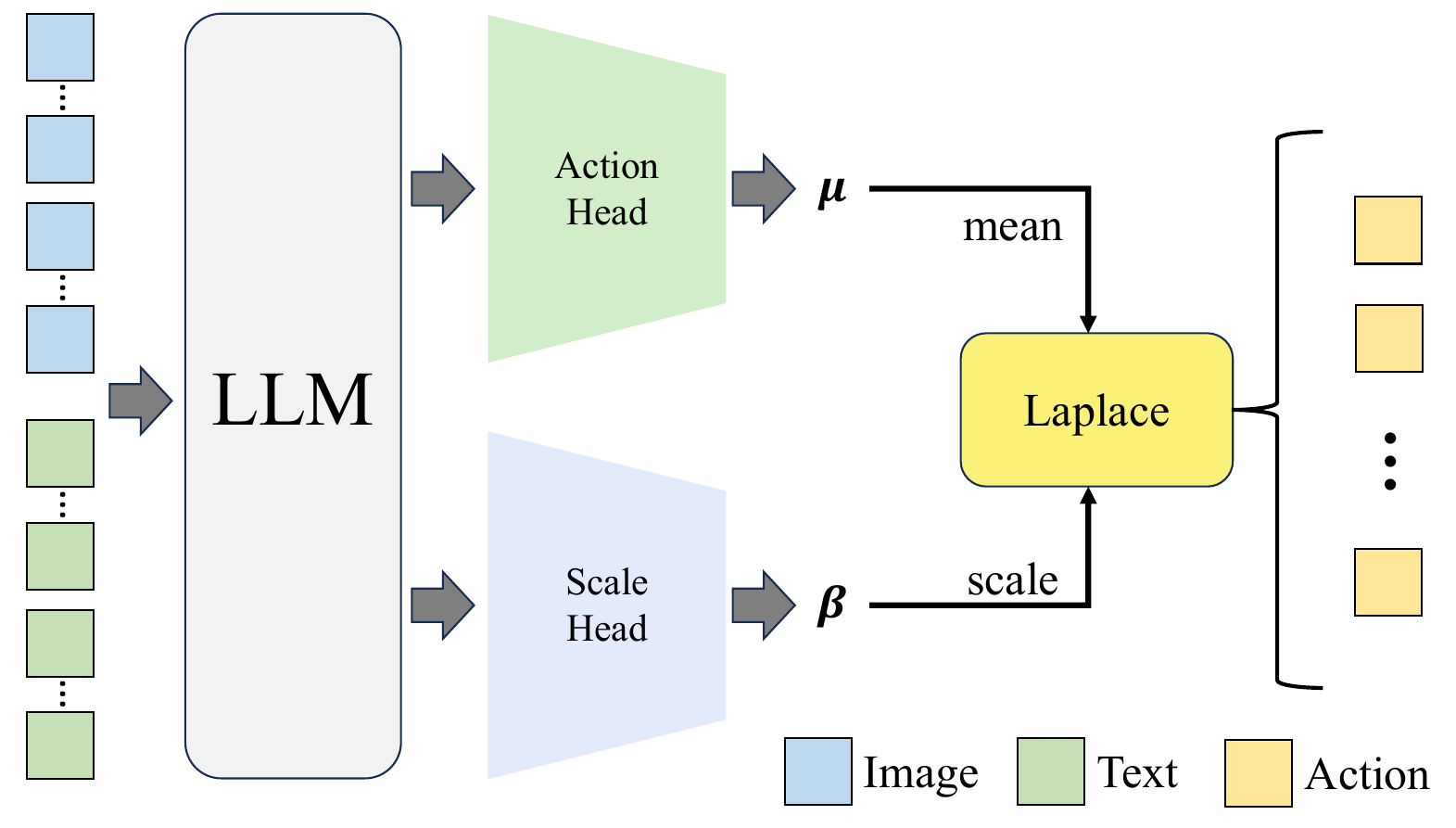} \\
  \caption{\textbf{Architecture for uncertainty-aware action generation.} The deterministic action output of the VLA policy is augmented with a parallel Laplace scale head to model action uncertainty.}
  \label{fig:net_scale}
\end{figure}

To model heteroscedastic uncertainty in action prediction, we introduce a scale head with the same MLP architecture as the action head, as shown in Figure~\ref{fig:net_scale}. This scale head outputs log-scale parameters $\boldsymbol{\beta}$ through:
\begin{equation}
    \boldsymbol{\beta} = \text{MLP}_{\text{scale}}(h),
\end{equation}
and is trained with negative log-likelihood (NLL) loss under a Laplace distribution assumption:
\begin{equation}
    \mathcal{L}_{\text{NLL}} = 
    \underbrace{|\mathbf{a}_{\text{gt}} - \boldsymbol{\mu}| \cdot e^{-\boldsymbol{\beta}}}_{\text{Data fit}} 
    + \underbrace{\boldsymbol{\beta}}_{\text{Uncertainty penalty}} + \log 2.
\end{equation}
The scale head is trained using a batch size of 8 and a learning rate of $5\times 10^{-4}$ over $1,000$ training iterations.


\subsection{Details of Reward Head}
\begin{figure}[t]
\begin{center}
\includegraphics[width=0.6\linewidth]{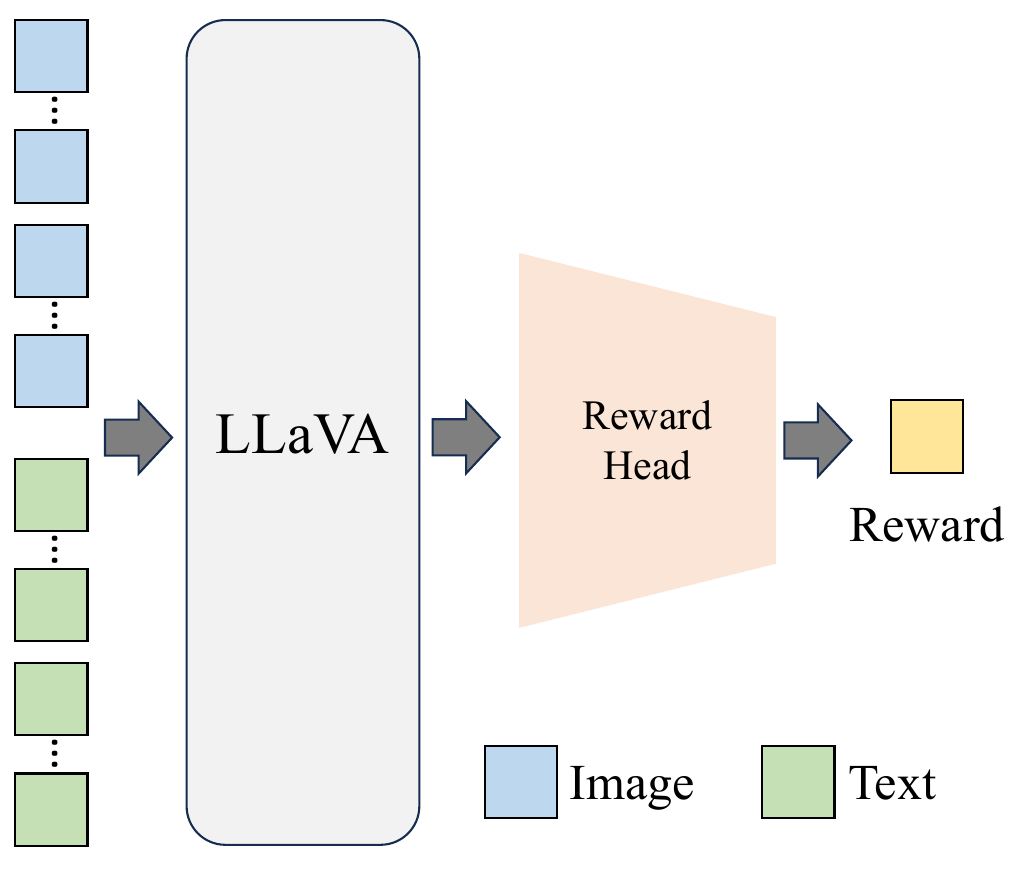}
\end{center}
\caption{\textbf{Network architecture of instant reflector.}}
\label{fig:rm}
\end{figure}

Our VLM-guided instant reflector integrates a pretrained vision-language model (LLaVA~\citep{liu2023llava}) with a lightweight reward head that predicts continuous reward signals, see Figure~\ref{fig:rm} for an overview. The VLM backbone is kept frozen to preserve its semantic capabilities, and only the reward head is trained. Given a video sequence $\{f_1, \dots, f_N\}$ generated by the world simulator, we uniformly sample 32 frames as visual input. The language prompt is formatted as: \textit{``Watch the video and determine whether it completes the task: $\{\mathbf{g}\}$ — answer only `Yes' or `No'.''} The VLM processes this input and extracts a pooled embedding, which is projected by the reward head to a scalar. A sigmoid activation yields a continuous reward $R \in [0,1]$, interpreted as the task completion probability. The reward head is trained with binary cross-entropy loss, using a batch size of 8, learning rate $1 \times 10^{-4}$, Adam optimizer, and 50 epochs, with input frames center-cropped to 384$\times$384 resolution.

\begin{table}[h]
\centering
\caption{\textbf{Performance of the instant reflector under different $\eta$.} The selected value 
$\eta=0.5$ achieves a balanced trade-off across accuracy, precision, recall, and F1-score.}
\resizebox{\linewidth}{!}
{
\begin{tabular}{l c c c c}
\toprule[1pt]
\multicolumn{1}{l}{$\eta$}& \multicolumn{1}{c}{Accuracy}& \multicolumn{1}{c}{Precision}& \multicolumn{1}{c}{Recall}& \multicolumn{1}{c}{F1-score}\\ 
\midrule
0.2&0.825&0.667&1.0&0.8\\
0.4&0.875&0.764&0.928&0.838\\
0.5 (Ours)&0.925&0.867&0.928&0.896\\
0.6&0.875&0.846&0.785&0.815\\
0.8&0.85&0.9&0.642&0.75\\
\bottomrule[1pt]
\end{tabular}
}

\label{table:reward_head_result}
\end{table}

\section{Analysis of Instant Reflector}
To investigate the sensitivity of model performance to the hyperparameter $\eta$, we conduct a series of experiments by varying $\eta$ over a predefined range (e.g., {0.2, 0.4, 0.5, 0.6, 0.8}). As shown in Table~\ref{table:reward_head_result}, we report four standard evaluation metrics: accuracy, precision, recall and F1-score. The results indicate that $\eta=0.5$ yields consistently strong performance across all metrics. 

\section{Analysis of World Simulator}

\subsection{Data Analysis and Distribution}
\begin{figure*}[htbp]
    \centering
    \begin{subfigure}[b]{0.32\textwidth}
        \centering
        \includegraphics[width=\textwidth]{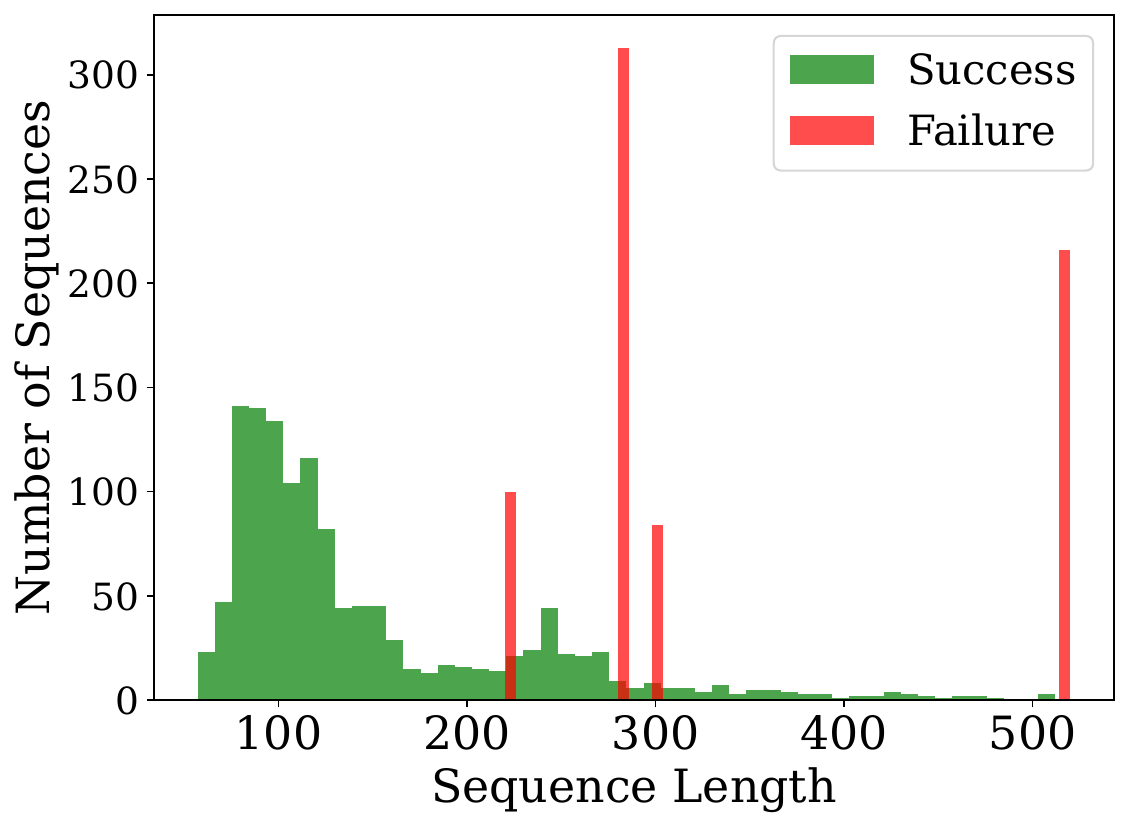}
        \caption{Length Distribution}
        \label{fig:outcom_com}
    \end{subfigure}
    \hfill
    \begin{subfigure}[b]{0.32\textwidth}
        \centering
        \includegraphics[width=\textwidth]{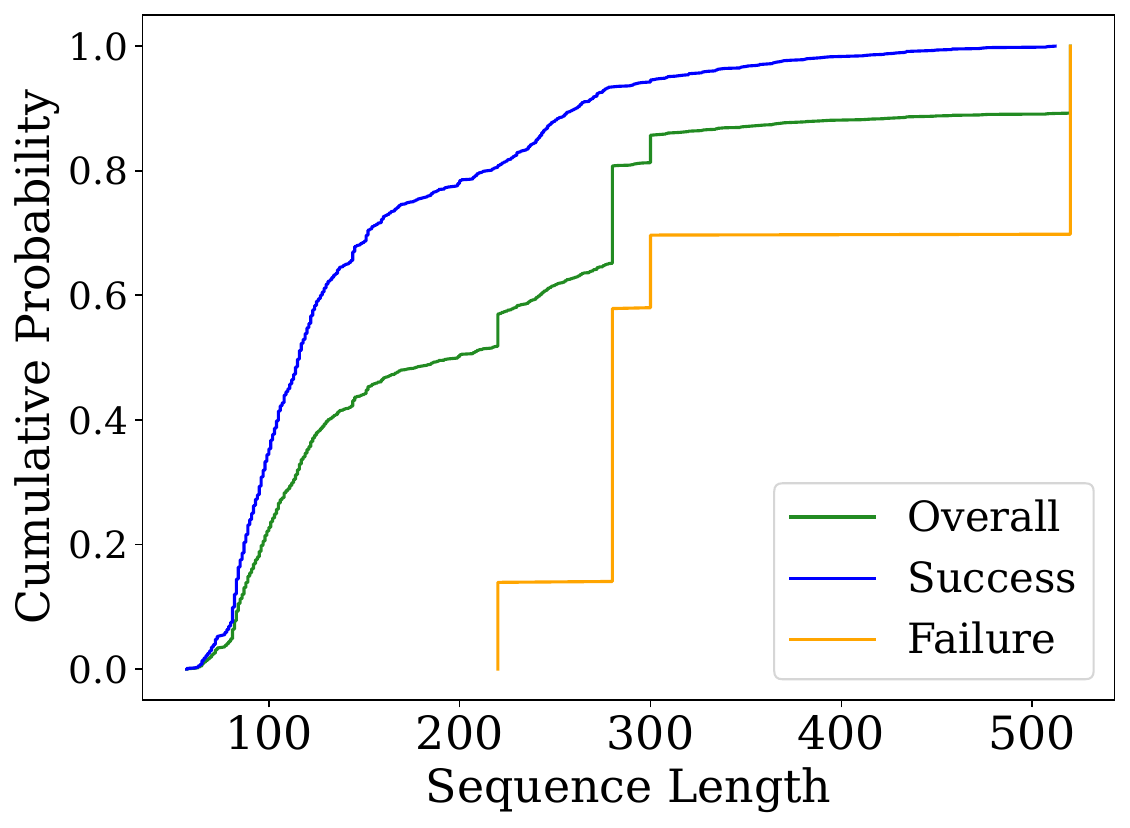}
        \caption{Cumulative Distribution Func.}
        \label{fig:cumu_dis}
    \end{subfigure}
    \hfill
    \begin{subfigure}[b]{0.32\textwidth}
        \centering
        \includegraphics[width=\textwidth]{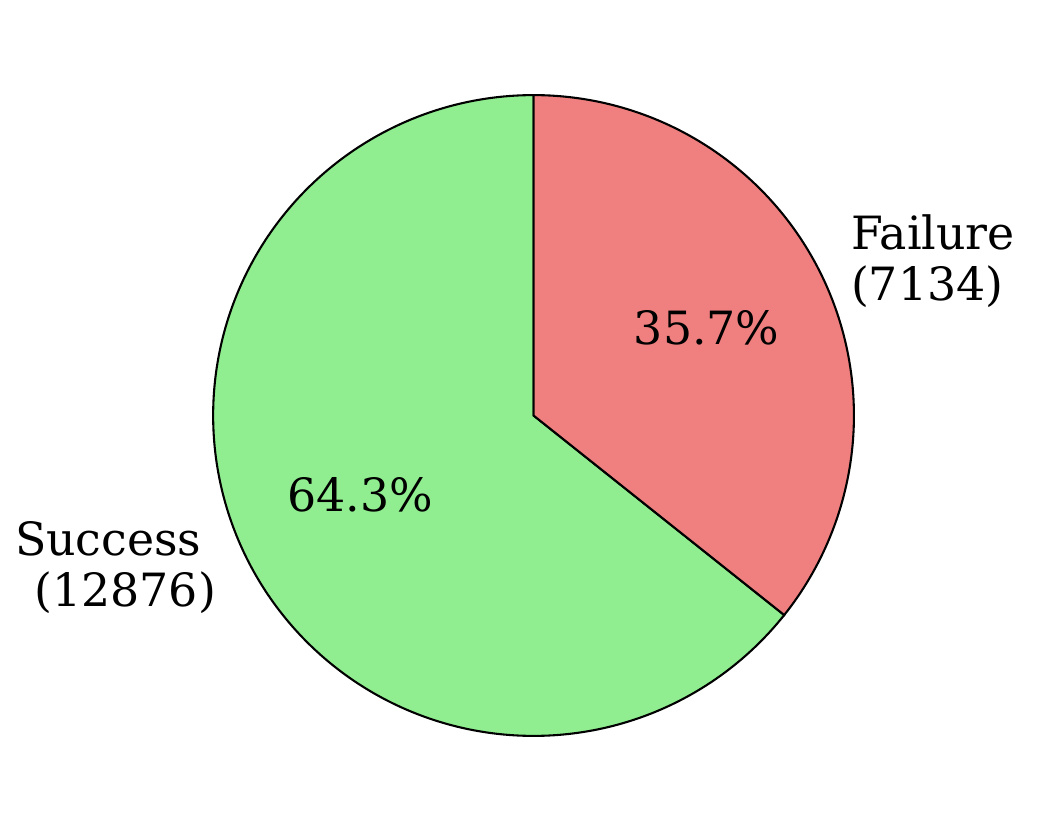}
        \caption{Task Outcome Proportion}
        \label{fig:task_outcome}
    \end{subfigure}
    \caption{\textbf{Training data analysis and distribution.}}
    \label{fig:data_distribution}
\end{figure*}

We provide a statistical analysis of the training data for the world simulator and instant reflector in Figure~\ref{fig:data_distribution}, including: (a) length distributions for successful vs. failed trajectories, (b) cumulative distribution functions by outcome, and (c) task outcome proportions. The bimodal distribution in successful trajectories motivated our dynamic termination mechanism, while the long-tailed length distribution informed our curriculum sampling strategy.

\subsection{Ablation Studies}
We evaluate the impact of training data, our proposed geometry-aware VGGT feature injection, and alternative representations (such as SAM and DINO) on the performance of the world simulator. Quantitative results are presented in Table~\ref{table:com_wm}, where ``w/o extra'' denotes the model trained without additional data, and ``w/o VGGT'' refers to the variant without our geometry-aware feature injection strategy. We evaluate using standard metrics: FID~\cite{NIPS2017FID}, FVD~\cite{unterthiner2019fvd}, PSNR~\cite{2008psnr}, SSIM~\cite{ssim}, and LPIPS~\cite{Zhang_2018_lpips}. As shown in Table~\ref{table:com_wm}, both the expanded training data and the VGGT latent features significantly contribute to building a more robust and physically consistent world model. Additional, we perturb the world model's outputs with: (i) Gaussian noise (mean = 0, variance = 0.1), and (ii) color perturbation, randomly adjusting brightness ($\pm20\%$), contrast ($\pm20\%$), saturation ($\pm20\%$), and hue ($\pm0.1$). As shown in Table~\ref{table:perturb}, thanks to the powerful pretrained VLM and our data augmentation, these perturbations only incur minor performance drop, validating our robustness to imperfect early-stage predictions.

\begin{table}[h]
\centering
\vspace{-3mm}
\resizebox{\linewidth}{!}
{
\begin{tabular}{l c c c}
\toprule[1pt]
\multicolumn{1}{l}{Task}& \multicolumn{1}{c}{Gaussian noise}& \multicolumn{1}{c}{Color perturbation}& \multicolumn{1}{c}{Original (Ours)}\\ 
\midrule
LIBERO-Spatial &85.4&87.0&87.6\\
\bottomrule[1pt]
\end{tabular}
}
\caption{\textbf{Impact of world model perturbations on final results}}
\vspace{-4.0mm}
\label{table:perturb}
\end{table}

\begin{table*}[h]
\centering
\caption{\textbf{Quantitative comparison of world models.}}
\resizebox{0.6\linewidth}{!}
{
\begin{tabular}{l c c c c c}
\toprule[1pt]
\multicolumn{1}{l}{Method}& \multicolumn{1}{c}{FID$\downarrow$}& \multicolumn{1}{c}{FVD$\downarrow$}& \multicolumn{1}{c}{PSNR$\uparrow$}& \multicolumn{1}{c}{SSIM$\uparrow$}& \multicolumn{1}{c}{LPIPS$\downarrow$}\\ 
\midrule
w/o extra&100.9978&116.7792&20.1600&0.7450&0.1385\\
w/o VGGT&70.9232&114.5264&23.1841&0.8219 &0.0806\\
Ours with DINO &41.7398&74.9209&23.4069&0.8511& 0.0602\\
Ours with SAM &40.6276&74.0298&23.6022&0.8532& 0.0599\\
Ours (CLIP+VGGT)&\textbf{39.1941} &\textbf{73.5313}&\textbf{23.8343}&\textbf{0.8579}&\textbf{0.0562}\\
\bottomrule[1pt]
\end{tabular}
}

\label{table:com_wm}
\end{table*}


\begin{figure*}[htbp]
\begin{center}
\includegraphics[width=\textwidth]{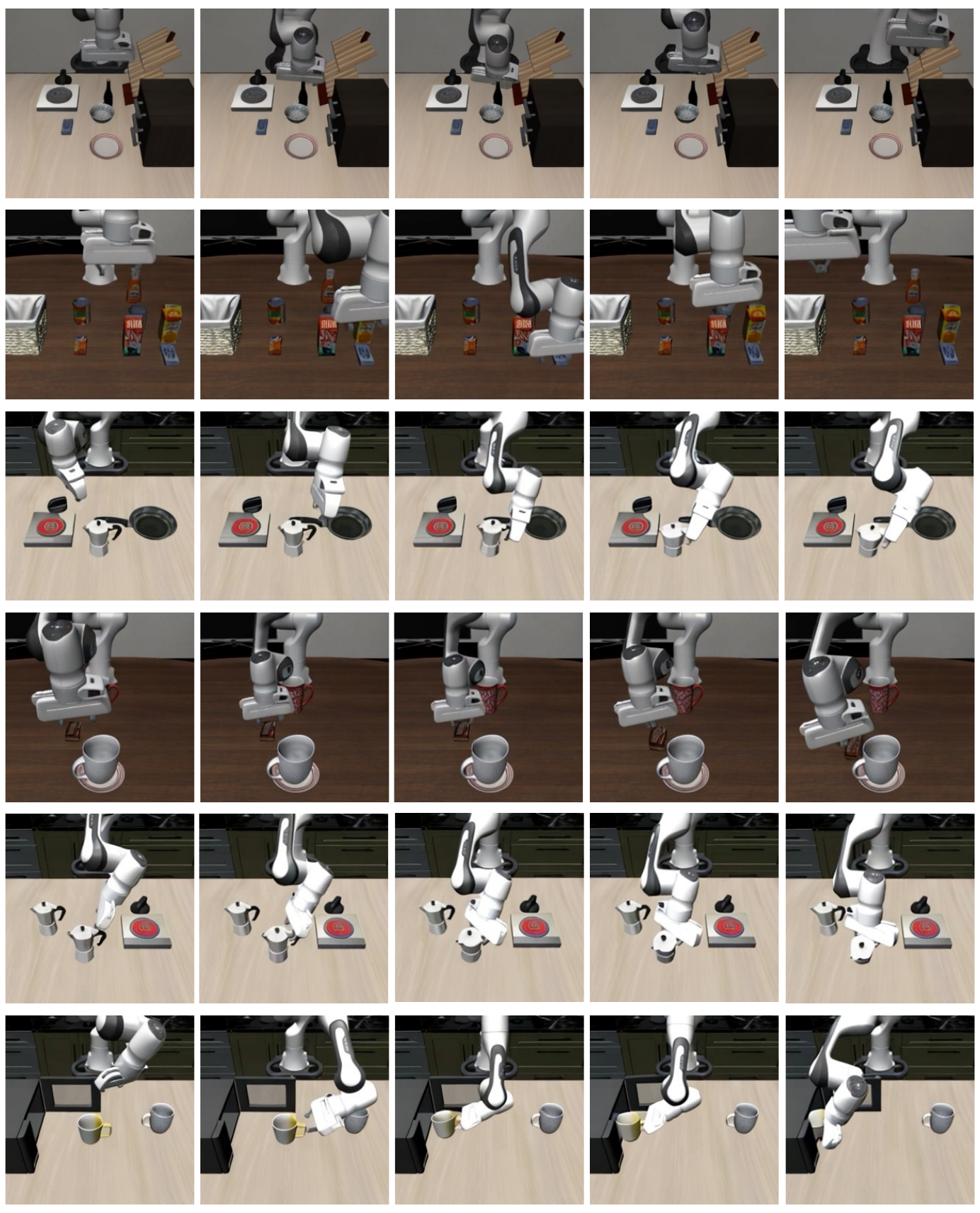}
\end{center}
\caption{\textbf{Failure trajectories synthesized by the world simulator.}}
\label{fig:wm_fail}
\end{figure*}

\begin{figure*}[htbp]
\begin{center}
\includegraphics[width=1\textwidth]{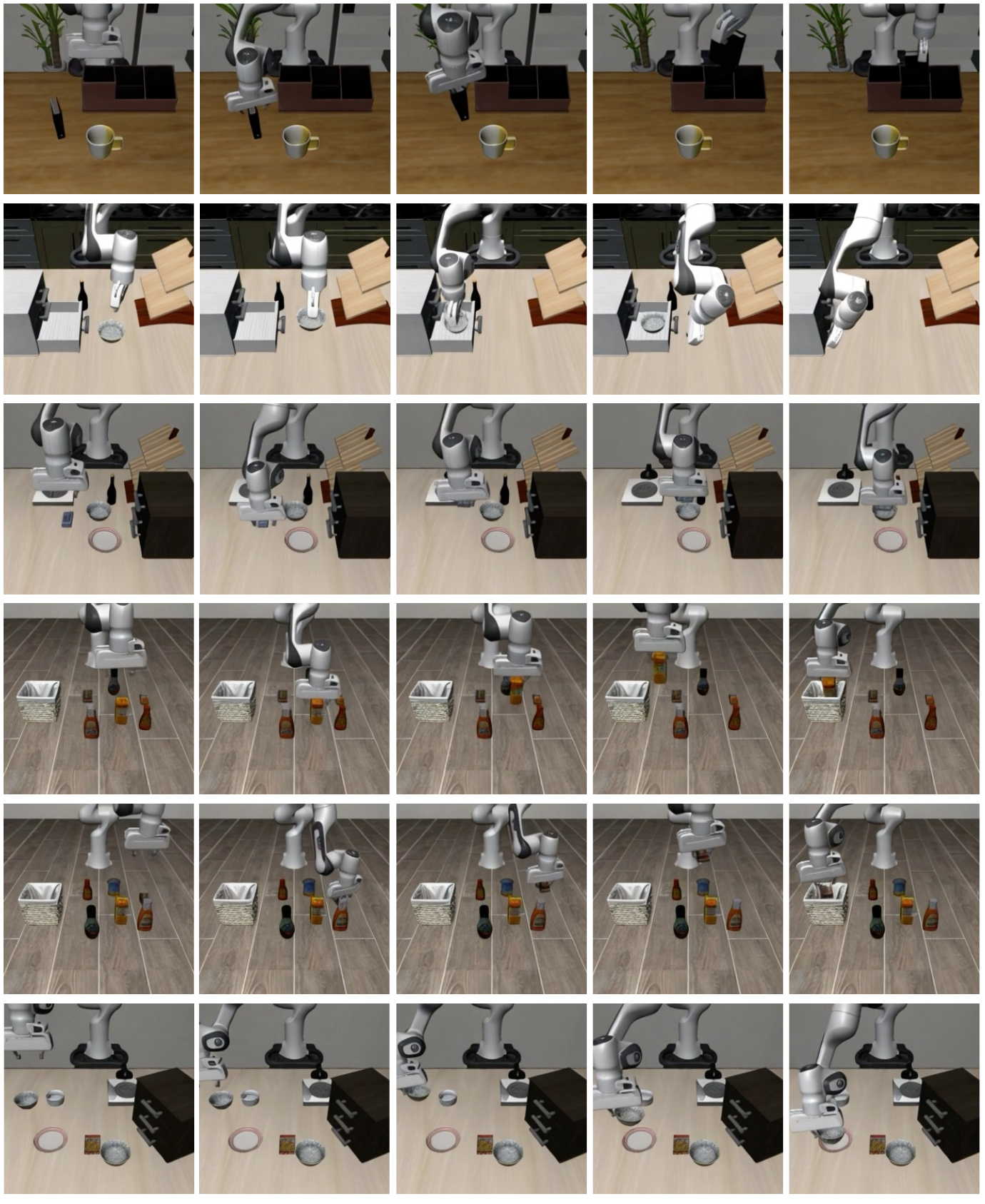}
\end{center}
\caption{\textbf{Success trajectories synthesized by the world simulator.}}
\label{fig:wm_success}
\end{figure*}

\subsection{More Results}
Figures~\ref{fig:wm_fail} and~\ref{fig:wm_success} show additional trajectories generated by the world simulator, demonstrating its ability to synthesize both successful and failed task executions.



{
    \small
    \bibliographystyle{ieeenat_fullname}
    \bibliography{main}
}


\end{document}